\documentclass{article}

\usepackage{microtype}
\usepackage{graphicx}
\usepackage{subfigure}
\usepackage{booktabs} 

\usepackage{hyperref}

\usepackage[accepted]{icml2023}

\usepackage{mathrsfs}
\usepackage{amsmath}
\usepackage{amssymb}
\usepackage{mathtools}
\usepackage{amsthm}

\DeclareMathOperator*{\argmin}{\arg\!\min}

\DeclareMathOperator{\Tr}{Tr}

\usepackage[capitalize,noabbrev]{cleveref}

\theoremstyle{plain}
\newtheorem{theorem}{Theorem}[section]

\newtheorem{lemma}[theorem]{Lemma}
\newtheorem{corollary}[theorem]{Corollary}
\theoremstyle{definition}
\newtheorem{definition}[theorem]{Definition}

\theoremstyle{remark}

\usepackage[textsize=tiny]{todonotes}

\icmltitlerunning{Sample Complexity Bounds for Learning High-dimensional Simplices in Noisy Regimes}

\begin{document}

\twocolumn[
\icmltitle{Sample Complexity Bounds for\\
Learning High-dimensional Simplices in Noisy Regimes}



\icmlsetsymbol{equal}{*}

\begin{icmlauthorlist}
\icmlauthor{Amir Hossein Saberi}{yyy}
\icmlauthor{Amir Najafi}{sch}
\icmlauthor{Seyed Abolfazl  Motahari}{comp}
\icmlauthor{Babak H. Khalaj}{yyy}
\end{icmlauthorlist}

\icmlaffiliation{yyy}{Department of Electrical Engineering, Sharif University of Technology, Tehran, Iran}
\icmlaffiliation{comp}{Department of Computer Engineering, Sharif University of Technology, Tehran, Iran}
\icmlaffiliation{sch}{School of Mathematics, Institute for Research in Fundamental Sciences (IPM), P.O. Box: 19395-5746, Tehran, Iran.}

\icmlcorrespondingauthor{Amir Hossein Saberi}{saberi.sah@ee.sharif.edu}
\icmlcorrespondingauthor{Amir Najafi}{najafi@ipm.ir}
\icmlcorrespondingauthor{Seyed Abolfazl  Motahari}{motahari@sharif.edu}
\icmlcorrespondingauthor{Babak H. Khalaj}{khalaj@sharif.edu}

\icmlkeywords{Machine Learning, ICML}

\vskip 0.3in
]



\printAffiliationsAndNotice{}  
\begin{abstract}
In this paper, we find a sample complexity bound for learning a simplex from noisy samples. Assume a dataset of size $n$ is given which includes i.i.d. samples drawn from a uniform distribution over an unknown simplex in $\mathbb{R}^K$, where samples are assumed to be corrupted by a multi-variate additive Gaussian noise of an arbitrary magnitude. We prove the existence of an algorithm that with high probability outputs a simplex having a $\ell_2$ distance of at most $\varepsilon$ from the true simplex (for any $\varepsilon>0$). Also, we theoretically show that in order to achieve this bound, it is sufficient to have $n\ge\left(K^2/\varepsilon^2\right)e^{\Omega\left(K/\mathrm{SNR}^2\right)}$ samples, where $\mathrm{SNR}$ stands for the signal-to-noise ratio. This result solves an important open problem and shows as long as $\mathrm{SNR}\ge\Omega\left(K^{1/2}\right)$, the sample complexity of the noisy regime has the same order to that of the noiseless case. Our proofs are a combination of the so-called sample compression technique in \citep{ashtiani2018nearly}, mathematical tools from high-dimensional geometry, and Fourier analysis. In particular, we have proposed a general Fourier-based technique for recovery of a more general class of distribution families from additive Gaussian noise, which can be further used in a variety of other related problems.
\end{abstract}

\section{Introduction}
\label{sec:intro}
Many practical problems in machine learning and data science can be naturally modeled by learning of high-dimensional geometric shapes from a given set of unlabeled data points. In particular, learning of simplices from randomly scattered points arises in many fields ranging from bioinformatics to remote sensing \citep{chan2009convex, schwartz2010applying, satas2017tumor}. A simplex is defined as the set of all convex combinations of $K+1$ points in $\mathbb{R}^K$, for $K\in\mathbb{N}$. Formally speaking, our problem can be stated as follows: assume $n$ i.i.d. samples are drawn from a uniform measure over an unknown simplex in $\mathbb{R}^K$. Also, each sample is assumed to be corrupted by an additive multi-variate white Gaussian noise with a covariance of $\sigma^2\boldsymbol{I}_{K\times K}$, for an unknown $\sigma$. In this regard, the main question that we try to address in this paper is: How large $n$ needs to be in terms of parameters such as $K$, $\sigma$, $\mathrm{SNR}$ and etc., such that the true simplex can be consistently approximated with an arbitrarily high probability? 

Learning of simplices is a well-studied task. Due to its practical importance, many heuristic algorithms are proposed to solve this problem in different scenarios such as noisy and sparse cases. In this paper, we focus on the theoretical aspects of this research area which still contains several open problems. There exists at least one efficient algorithm for learning of simplices that comes with a theoretical guarantee (see Section \ref{sec:intro:LR}), but still suffers from a very large sample complexity, e.g., it requires $n\geq\Omega\left(K^{22}\right)$. In a concurrent line of work, researchers work on deriving information-theoretic bounds on the sample complexity of this problem irrespective of their time and memory complexity. Recently, optimal information-theoretic sample complexity bounds are derived for the noiseless case where Maximum Likelihood Estimator (MLE) is used. The output of MLE for this problem is the minimum volume simplex which contains all the samples which runs in exponential time\footnote{Finding a polynomial algorithm with provable guarantees that comes with a reasonable dependency on $K$ is still an open problem.}. \citet{najafi2021statistical} proved that MLE is a PAC-learning algorithm for finding a $K$-simplex with an asymptotically decreasing Total Variation (TV) distance from the true one. They showed that a maximum of $\tilde{\Omega}\left(K^2/\varepsilon \right)$ (Logarithmic terms are ignored) samples are sufficient to estimate the true simplex up to a TV distance of $\varepsilon>0$. However, the noiseless case is unrealistic since we do not have access to clean data in real-world situations. On the other hand, the minimum-volume inclusive simplex which is proposed by \citet{najafi2021statistical} is no longer a valid solution when samples are noisy, since the corrupted samples are not necessarily contained inside the true simplex. 

In this work, we aim at finding the information-theoretic sample complexity of learning simplices in noisy regimes. Mathematically speaking, we assume samples are generated according to the following equation:
\begin{equation}
\boldsymbol{y}_i = \boldsymbol{V} \boldsymbol{\phi}_i + \boldsymbol{z}_i \quad,\quad i=1,\ldots,n,
\end{equation}
where $\boldsymbol{V}$ is a $K\times (K+1)$ matrix that includes the vertices of the true simplex $\mathcal{S}_T$ as its columns. Each $\boldsymbol{\phi}_i \in \mathbb{R}^K$ is drawn from a uniform Dirichlet distribution and $\boldsymbol{z}_i \in \mathbb{R}^K$ are sampled from a multivariate Gaussian distribution $\mathcal{N}\left(0,\sigma^2\boldsymbol{I}_K\right)$. We tackle the task of estimating the true simplex via an exponential-time estimator $\hat{\mathcal{S}}=\hat{\mathcal{S}}\left(\boldsymbol{y}_{1:n}\right)$. In particular, we prove that having at most $\tilde{\Omega}\left(K^2/\varepsilon^2 \right)e^{\Omega\left(K/\mathrm{SNR}^2\right)}$ noisy samples from a $K$-simplex is sufficient to estimate the simplex with a $\ell_2$-distance of at most $\varepsilon$, for any $\epsilon>0$. Here, $\mathrm{SNR}$ denotes the ratio of the signal, i.e., maximum standard deviation of the uniform measure over $\mathcal{S}_T$, to the standard deviation of the noise per component, i.e., $\sigma$. This result solves one of the main open problems in this area and could justify an interesting previously observed phenomenon in practice: Almost all heuristic algorithms perform close the noiseless case as long as $\mathrm{SNR}$ is larger than a presumably sub-linear function of dimension $K$. However, they experience a sudden decline in performance as $\mathrm{SNR}$ is decreased further from that point.

Our approach to prove the above-mentioned bound has four main steps. 1) We find a ball in $\mathbb{R}^K$ which with high probability contains the true simplex. 2) This ball is then quantized through covering with enough isolated points. Each $K+1$ combination of such points forms a candidate simplex, and consequently a candidate density to approximate the true simplex. As long as the covering is accomplished with a sufficiently high precision, we prove the existence of at least one candidate that has a small TV-distance from the main simplex. 3) In the next step, we choose a noise-corrupted density, i.e., the convolution of a uniform measure over a candidate simplex with Gaussian distribution, from the candidate set such that, with high probability, it has a small TV distance from that of the true simplex. This can be done using some aspects of the so-called ``sample compression" technique. 4) In the final step, we propose a novel Fourier-based technique to show that as long as the noise-smoothed versions of any two simplices are close to each other in TV-distance, their underlying simplices are also close to each other in the sense of $\ell_2$-distance.


\subsection{Related Works}
\label{sec:intro:LR}
We categorize the existing works based on their focus, which could be either the efficiency of the proposed algorithms, or the fundamental sample complexity of the problem irrespective of its time complexity. In the former, the main purpose is mostly to provide a heuristic solution with promising results in practice. However, papers in the latter category aim to find fundamental and information-theoretic limitations for the problem. In the remainder of this part, we review a number of works from both categories.

\citet{anderson2013efficient} proved that by having $\Omega\left(K^{22}\right)$ noiseless samples, one can estimate the true simplex via an polynomial-time algorithm. The core idea is to utilize the third moment and local search techniques from Independent Component Analysis (ICA) research. To the best of our knowledge, finding polynomial-time algorithms with a more relaxed dependence to $K$ is still an open problem. Concurrently, \citet{najafi2021statistical} proved that sample complexity of the MLE in noiseless case is $\tilde{\Omega}\left(K^2/\varepsilon\right)$, where $\varepsilon$ is the permissible TV-distance of the output of the algorithm from the true simplex. Since MLE in this case runs in exponential time, they also provide an alternative heuristic approach as a practical surrogate to MLE, which still does not come with a rigid theoretical guarantee.

Theoretical attempts to tackle this problem in the noisy setting is limited to the work of \citet{bhattacharyya2020near}. Their work is also based on the sample compression technique, originally used in the seminal work of \citet{ashtiani2018nearly} in learning high-dimensional Gaussian mixture models. \citet{bhattacharyya2020near} prove that a noisy simplex can be learned using $\Omega\left(K^2\right)$ samples only if there exists at least one sample near each vertex of the simplex. In fact, this is a very strong assumption, and a back of the envelope calculation shows that one needs around $\tilde{\Omega}\left((1/\varepsilon)^{K}\right)$ samples to guarantee the occurrence of this event. 

From a practical point of view, several heuristic methods have been introduced so far in order to deal with real-world problems in bioinformatics, hyper-spectral imaging and etc \citep{piper2004object, bioucas2012hyperspectral, lin2013endmember}.
In hyper-spectral imaging, one aims at finding the distribution of the constituent elements in an area by examining remote hyper-spectral images. Each pixel in such an image can be thought as a random convex combination of a finite set of fixed prototypes which correspond to pure elements that exist in that region. Therefore, the problem of finding constituent minerals would natural translate into estimating an unknown simplex from a set of presumably uniform samples \citep{ambikapathi2011chance, agathos2014robust, zhang2017robust}. Learning of simplices in bioinformatics usually arises in the study of complex tissues. A complex tissue is composed of multiple cell-types --a group of cells with similar characteristics-- such as blood, brain, or even tumor cells \citep{tolliver2010robust,zuckerman2013self}. Bulk data from complex tissues, such as gene expression level vectors, can again be modeled by convex combinations of its constituent cell-types. In this line of work, researchers aim at finding the structure of tissues through studying the state of cell-types, which again leads into learning a high-dimensional simplex \citep{shoval2012evolutionary, korem2015geometry}. 


The rest of the paper is organized as follows: In Section \ref{sec:notation}, we formally define the problem and present our notation and definitions. Our main theoretical results as well as the algorithm that achieves our bounds are discussed in Section \ref{sec:main}. Finally, Section \ref{sec:conc} concludes the paper and presents some suggestions for future works.


\section{Notation and Definitions}
\label{sec:notation}

We use the same notations as \citep{najafi2021statistical}. A $K$-simplex is defined as the set of all convex combination of $K+1$ affinely independent points in $\mathbb{R}^{K}$. Let $\boldsymbol{V}= \left[\boldsymbol{v}_0\vert\boldsymbol{v}_1\vert\cdots\vert \boldsymbol{v}_{k}\right]\in\mathbb{R}^{K\times(K+1)}$ be a matrix whose columns represent vertices of the simplex, then $K$-simplex $\mathcal{S}$ can be defined as
\begin{equation*}
\mathcal{S} = \left\{ \boldsymbol{V}\boldsymbol{\phi} \bigg\vert~  \boldsymbol{\phi} \in \mathbb{R}^{K+1} ,~ \boldsymbol{\phi} \succ \boldsymbol{0},~ \boldsymbol{\phi}^{T}\boldsymbol{1}  = 1  \right\}.
\end{equation*}
Also, let us  denote the set of all possible $K$-simplices in $\mathbb{R}^K$ by $\mathbb{S}_{K}$. We denote the uniform probability measure over a simplex $\mathcal{S}$ by $\mathbb{P}_{\mathcal{S}}$, and its probability density function by $f_\mathcal{S}\left(x\right)$. Thus, $f_\mathcal{S}\left(x\right)$ can be written as
\begin{equation*}
f_\mathcal{S}\left(x\right) = \frac{\boldsymbol{1}\left(\boldsymbol{x} \in \mathcal{S}\right)}{\mathrm{Vol}\left(\mathcal{S}\right)},
\end{equation*}
where $\mathrm{Vol}\left(\mathcal{S}\right)$ denotes the Lebesgue measure (or volume) of $\mathcal{S}$. 

The noisy simplex family, i.e., the class of distributions formed by {\it {convolving}} uniform probability density function over simplices in $\mathbb{S}_K$ by $G_{\sigma}\triangleq\mathcal{N}\left(\boldsymbol{0},\sigma^2\boldsymbol{I}\right)$, is denoted by $\mathbb{G}_{K,\sigma}$. Mathematically speaking,
\begin{equation}
\mathbb{G}_{K,\sigma}
\triangleq
\left\{
f_{\mathcal{S}}*G_{\sigma}
\vert~
\mathcal{S}\in\mathbb{S}_K
\right\},
\end{equation}
where $*$ denotes the convolution operator. Obviously, the distribution of input data points $\boldsymbol{y}_i$ lie in $\mathbb{G}_{K,\sigma}$. With a little abuse of notation, we use the term ``class of simplices" to both refer to $\mathbb{S}_K$ and also $\left\{f_{\mathcal{S}}\vert~\mathcal{S}\in\mathbb{S}_K\right\}$ whenever the difference is clear from the context. In a similar fashion, we refer to $\mathbb{G}_{K,\sigma}$ as the class of ``noisy simplices".

In order to measure the difference between two distributions, we use both $\ell_2$ and total variation distance. Consider two probability measures $\mathbb{P}_1$ and $\mathbb{P}_2$, with respective density functions $f_1$ and $f_2$, which are defined over $\mathbb{R}^K$. Then, TV distance between $\mathbb{P}_1$ and $\mathbb{P}_2$ can be defined as
\begin{align*}
\operatorname{TV}(\mathbb{P}_1, \mathbb{P}_2) \triangleq & \sup_{\mathrm{A} \in \mathcal{B}}{|\mathbb{P}_1\left(\mathrm{A}\right) - \mathbb{P}_2\left(\mathrm{A}\right)|} 
=  \frac{1}{2}\|f_1 - f_2\| _1,
\end{align*}
where $\mathcal{B}$ is the set of all Borel sets in $\mathbb{R}^K$.

\begin{definition}[PAC-Learnability of a distribution class in realizable setting]
\label{definition:PAC}
We say a class of distributions $\mathcal{F}$ is PAC learnable in realizable setting, if there exists a learning method which for any distribution $g \in \mathcal{F}$ and any $\epsilon,\delta>0$, outputs an estimator $\hat{g}$ using $n \geq \mathrm{poly}\left(1/\epsilon, 1/\delta \right)$ i.i.d. samples from $g$, which with probability at least $1- \delta$ satisfies
\begin{equation}
\label{eq: realizable pac learning}
\|\hat{g} - g\|_{\mathrm{TV}} \leq \epsilon.
\end{equation}
\end{definition}
We also need to define a series of geometric restrictions for the simplex in noisy cases. In fact, those simplices that are significantly stretched toward on particular direction can be shown to be more prune to noise than those with some minimum levels of geometric regularity. We discuss the necessity of such definitions in later stages. Similar to \citet{najafi2021statistical}, let us denote the volume of the largest facet of a $K$-simplex (here, volume needs be calculated in $\mathbb{R}^{K-1}$) by $\mathcal{A}_{\max}\left(\mathcal{S}\right)$, and the length of the largest line segment inside the simplex (diameter) by $\mathcal{L}_{\max}\left(\mathcal{S}\right)$. In this regard, we define the isoperimetricity of a $K$-simplex as follow:
\begin{definition}[$\left(\underline{\theta},\bar{\theta}\right)$-isoperimetricity of simplices]
A $K$-simplex $\mathcal{S}\in\mathbb{S}_K$ is defined to be $\left(\underline{\theta},\bar{\theta}\right)$-isoperimetric if the following inequalities hold:
\begin{align*}
\mathcal{A}_{\max}\left(\mathcal{S}\right)
~\leq~
\bar{\theta} \mathrm{Vol}\left(\mathcal{S}\right)^{\frac{K-1}{K}},
\\
\mathcal{L}_{\max}\left(\mathcal{S}\right)
~\leq~
\underline{\theta}K \mathrm{Vol}\left(\mathcal{S}\right)^{\frac{1}{K}}.
\end{align*}
\end{definition}
The overall concept of isoperimetricity for simplices reflects the fact that for a simplex to be (even partially) recoverable from noisy data, it should not be stretched too much in any direction or having highly acute angles. In other words, sample complexity in noisy regimes is also affected by the geometric shape of the underlying simplex.

\begin{definition}[$\epsilon$-representative set]
For any $\epsilon>0$, we say that a finite set of distributions $\mathcal{G}$ is an $\epsilon$-representative set for a distribution class $\mathcal{F}$, if for any distribution $f\in \mathcal{F}$, there exists at least one $g \in \mathcal{G}$ that satisfies
\begin{equation*}
\|f-g\|_{\mathrm{TV}} \leq \epsilon.
\end{equation*}
\end{definition}

Throughout the paper, we use bold lowercase letters to show scalars, bold letters to show vectors and bold uppercase letters to show matrices. We also use light uppercase letters to show random variables and light lower case letters to show realization of a random variable. 
In this paper and for the sake of simplicity in notation, for any fixed $\bar{\theta},\underline{\theta}>0$, whenever we say the class of simplices we actually mean the class of $\left(\underline{\theta},\bar{\theta}\right)$-isoperimetric simplices.

\subsection{Problem Definition}

In the first three parts of the paper, we aim at proving the PAC-learnability of noisy simplices family $\mathbb{G}_{K,\sigma}$. The final part is devoted to showing that if two members in $\mathbb{G}_{K,\sigma}$ are close to each other, e.g., the true noisy simplex and its consistent estimator, then the simplices are also consistently close.

Based on definition \ref{definition:PAC}, and in order to show that the class of noisy $K$-simplices is PAC-learnable, one should find an algorithm that for all $f_{\mathcal{S}}*G_{\sigma} \in \mathbb{G}_{K,\sigma}$, any positive $\varepsilon,\delta>0$, and given a dataset $\boldsymbol{D} = \{\boldsymbol{y}_1, \boldsymbol{y}_2, \cdots, \boldsymbol{y}_n \}$ of i.i.d. samples drawn from $f_{\mathcal{S}}*G_{\sigma} \in \mathbb{G}_{K,\sigma}$, as long as $n  \geq \mathrm{poly}\left(1/\varepsilon, 1/\delta \right)$, outputs a noisy simplex $f_{\widehat{\mathcal{S}}}*G_{\sigma}$ such that
\begin{equation}
\mathbb{P}\left[
\|
\left(f_{\widehat{\mathcal{S}}} - f_{\mathcal{S}}\right)*G_{\sigma}
\|_{\mathrm{TV}} 
\geq \varepsilon\right] \leq \delta.
\end{equation}
This is due to the fact that we assume samples to be drawn from $\mathbb{P}_{\mathcal{S}}$ and then are corrupted by an additive independent zero mean Gaussian noise with a covariance matrix of $\sigma^2\boldsymbol{I}$. Our ultimate goal in the first three parts of the paper is the following: To derive explicit polynomial forms for the lower-bound $n\ge\mathrm{poly}\left(1/\varepsilon,\log\left(1/\delta\right)\right)$, which (as we show in Theorem \ref{Theorem2}) turns out to be
$$
n\ge \tilde{\Omega}\left[\frac{K^2}{\varepsilon^2}\log\frac{1}{\delta}\right].
$$
The final part is dedicated to show that as long as (with high probability)
$
\left(f_{\widehat{\mathcal{S}}} - f_{\mathcal{S}}\right)*G_{\sigma}\leq\varepsilon,
$
then we also have
$$
\left\Vert 
f_{\widehat{\mathcal{S}}} - f_{\mathcal{S}}
\right\Vert_2\leq
\varepsilon e^{\Omega\left(\frac{K}{\mathrm{SNR}^2}\right)},
$$
where $\mathrm{SNR}\triangleq \left(\mathcal{L}_{\max}\left(\mathcal{S}\right)/K\right)/\sigma$ denotes the signal-to-noise ratio. It can be easily checked that based on the definition of a uniform Dirichlet distribution, $\mathcal{L}_{\max}/K$ represents the maximum component-wise standard deviation associated to $f_{\mathcal{S}}$. This will prove our claims in Section \ref{sec:intro}.


\section{Statistical Learning of Noisy Simplices}
\label{sec:main}
Before going through the details, let us first present an sketch of proof for PAC-learnability of noisy simplices. 

({\it {Bounding the candidate set}}): In order to estimate the true simplex from noisy i.i.d. samples, we first split our data in half and use the first half to restrict the set of all $K$-simplices in $\mathbb{S}_{K}$ to a bounded set $\mathbb{S}^{\mathcal{D}}_{K}$ which consists of all the simplices that happen to entirely fall within a finite ball. This way, we can eliminate very far candidates and thus focus on simplices that are placed near the data samples. In this regard, we construct a bounded version of $\mathbb{S}_K$, denoted by $\mathbb{S}^{\mathcal{D}}_{K}$, and prove that it includes (with high probability) the true underlying simplex. 

({\it {Quantization}}): We quantize this bounded set and create a finite $\epsilon$-representative set of $K$-simplices denoted by $\widehat{\mathbb{S}}^{\mathcal{D}}_{K} = \{\mathcal{S}_1, \mathcal{S}_2, \cdots, \mathcal{S}_M\}$ for $M\in\mathbb{N}$, such that for each simplex  $\mathcal{S} \in \mathbb{S}^{\mathcal{D}}_{K}$, there exists some $i\in\{1,2,\cdots,M\}$ where $\|\mathbb{P}_{\mathcal{S}_i}- \mathbb{P}_{\mathcal{S}}\|_{TV} \leq \epsilon$. 

({\it {Density selection}}): In this part, we use the second half of data and try to choose the best simplex in $\widehat{\mathbb{S}}^{\mathcal{D}}_{K}$. By the best simplex, we mean the one with the minimum TV-distance from the true simplex. We show that as long as the above-mentioned sample complexity bound is satisfied, the output of this selection procedure fall within a $\varepsilon$ TV-distance of $f_{\mathcal{S}_T}*G_{\sigma}$.

({\it {Denoising}}): Finally, we show that estimating a noisy version of a simplex leads to consistent estimation for the simplex as well. This completes our proof.

\subsection{Bounding The Candidate Set}
In this part, we show how the first half of the dataset can be utilized in order to bound the set of all candidate simplices into a ball with a finite radius in $\mathbb{R}^K$. This procedure is crucial for later stages of the proof. In this regard, first let us review the generation process of noisy samples $\boldsymbol{y}_i$ for $i\in[n]$:
\begin{align}
\boldsymbol{y} \sim f_{\mathcal{S}_T}*G_{\sigma} \implies & \boldsymbol{y} = \boldsymbol{x} + \boldsymbol{z},
\nonumber \\ &   \boldsymbol{x} = \boldsymbol{V}_{\mathcal{S}}\boldsymbol{\phi}, \quad \boldsymbol{\phi} \sim \mathrm{Dir}\left(1,1,\cdots,1\right),
\nonumber \\ & \boldsymbol{z} \sim \mathcal{N}\left(\boldsymbol{0}, \sigma \boldsymbol{\mathrm{I}}\right),
\label{noisy simplex definition}
\end{align}
where $\boldsymbol{V}_{\mathcal{S}}$ denotes the vertex matrix for $\mathcal{S}$.

The following lemma shows that having enough samples from a noisy simplex $f_{\mathcal{S}}*G_{\sigma}$, one can find a hyper-sphere in $\mathbb{R}^K$ which (with high probability) contains $\mathcal{S}$.
\begin{lemma}[Creating $\mathbb{S}^{\mathcal{D}}_K$]
\label{lemma2}
Suppose that we have a set of i.i.d. samples $\boldsymbol{D} = \left\{\boldsymbol{y}_1,\boldsymbol{y}_2,\cdots,\boldsymbol{y}_{2m}\right\}$ from $f_{\mathcal{S}}*G_{\sigma}$ for $m\in\mathbb{N}$. If $m \geq 1000(K+1)(K+2)\log{\frac{6}{\delta}}$, then the true simplex $\mathcal{S}$ with probability at least $1- \delta$ is confined in a $K$-dimensional sphere with center point $\boldsymbol{p}$ and radius $R$, where $R$ and $\boldsymbol{p}$ are defined as follows:
\begin{align}
\mathrm{R} =  8\sqrt{(K+1)(K+2)D}
\quad,\quad
\mathrm{\boldsymbol{p}} = \frac{1}{2m}\sum_{i=1}^{2m}{\boldsymbol{y}_i},
\end{align}
and the variance of the noise can be upper-bounded as $\sigma^2 \leq \frac{D}{K-2} = \mathrm{R}_n$, where $D$ is defined as
$$
\mathrm{D} = \frac{1}{2m}\sum_{i =1}^{m}{\|\boldsymbol{y}_{2i} - \boldsymbol{y}_{2i-1}\|_2^2}.
$$
\end{lemma}
Proof can be found in Section \ref{sec:proof of lemmas} of the Appendix. This way, the set $\mathbb{S}^{\mathcal{D}}_K$ can be fixed. Next, we discuss how to properly quantize this set in order to choose an appropriate final candidate for the true simplex, denoted by $\widehat{\mathcal{S}}$.

\subsection{Quantization}

Let us call the hyper-sphere of Lemma \ref{lemma2} as $\mathrm{C}^{K}(\boldsymbol{p}, R)$, which with high probability contains the true simplex. In this part, we choose a set of points $\mathrm{T}^l_\epsilon(\mathrm{C}^{K}(\boldsymbol{p}, R)) = \{\boldsymbol{p}_1, \boldsymbol{p}_2, \cdots,  \boldsymbol{p}_l\}$ inside this sphere such that for any point $\boldsymbol{x} \in \mathrm{C}^{K}(\boldsymbol{p}, R)$, there exists some $i\in\{1,2,\cdots,l\}$ such that $\|\boldsymbol{x} - \boldsymbol{p}_i\|_{2} \leq \epsilon$. We call $\mathrm{T}^l_\epsilon(\mathrm{C}^{K}(\boldsymbol{p}, R))$ a covering set for $\mathrm{C}^{K}(\boldsymbol{p}, R)$. One way to construct such covering set is to uniformly draw points from the sphere. We show that if the number of drawn random points exceeds $O\left(\left(1+\frac{2R}{\epsilon}\right)^{2K}\right)$, then with high probability, they form an $\epsilon$-covering set for $\mathrm{C}^{K}(\boldsymbol{p}, R)$. 

Suppose that we construct a random covering set as described above and denote it with $\mathrm{T}_{\epsilon}(\mathrm{C}^{K}(\boldsymbol{p}, R))$. Now, using each $K+1$ distinct points in $\mathrm{T}_\epsilon(\mathrm{C}^{K}(\boldsymbol{p}, R))$ we can build a $K$-simplex. Assume we collect all such simplices in a set called $\widehat{\mathbb{S}}(\mathrm{C}^{K}(\boldsymbol{p}, R))$, i.e.,
\begin{align*}
&\widehat{\mathbb{S}}(\mathrm{C}^{K}(\boldsymbol{p}, R)) =
\\ &\left\{\mathcal{S}(\boldsymbol{x}_1, \ldots, \boldsymbol{x}_{K+1}) \bigg\vert~ \boldsymbol{x}_i \in \mathrm{T}_{\epsilon}(\mathrm{C}^{K}(\boldsymbol{p}, R)), ~ i \in \left[K+1\right] \right\}.
\end{align*}
Obviously, there exist at most $\binom{\vert \mathrm{T}_{\epsilon}(\mathrm{C}^{K}(\boldsymbol{p}, R))\vert}{K+1}$ simplices in $\widehat{\mathbb{S}}(\mathrm{C}^{K}(\boldsymbol{p}, R))$. The following lemma states that this set is a sufficiently good representative for all $\left(\underline{\theta},\bar{\theta}\right)$-isoperimetric $K$-simplices inside $\mathrm{C}^{K}(\boldsymbol{p}, R)$.
\begin{lemma}
\label{quantization lemma}
For any $\epsilon \in \left(0,1\right)$, denote the set of all possible $K$-simplices with the vertices in $ \mathrm{T}_{\frac{\alpha \epsilon}{K+1}}(\mathrm{C}^{K}(\boldsymbol{p}, R))$ as $\widehat{\mathbb{S}}(\mathrm{C}^{K}(\boldsymbol{p}, R))$. Then, the resulting set is an $\epsilon$-representative set for all $\left(\underline{\theta},\bar{\theta}\right)$-isoperimetric $K$-simplices in $\mathrm{C}^{K}(\boldsymbol{p}, R)$, as long as we have:
\begin{equation*}
\alpha =  \frac{\mathrm{Vol}\left(\mathcal{S}\right)^{{1}/{K}}}{5\bar{\theta}}.
\end{equation*}
\end{lemma}
The proof can be found in Section \ref{sec:proof of lemmas} of the Appendix. This way, the finite candidate set $\mathbb{S}^{\mathcal{D}}_K$ can be formed and we can jump to the next stage of our algorithm for finding a ``good" candidate for the true simplex.

\subsection{Density Selection}

We take advantage of the fundamental result in \citet{devroye2012combinatorial} which plays a central role in the remainder of our derivations in this subsection. At this stage, we have already created a finite set of representative simplices $\widehat{\mathbb{S}}^{\mathcal{D}}_{K}$ and set out to find the ``best" simplex in this set. This can be done using the following theorem:
\begin{theorem}[Theorem 6.3 of \citep{devroye2012combinatorial}]
\label{combinatoeic algorithm}
Let $\mathcal{F}$ be a finite class of distributions consisting of $M$ distinct members $\{ f_{1}, f_2, \cdots,f_M \}$. Also, suppose we have $n \geq \frac{\log{(3M^2/\delta)}}{2\epsilon^2}$ i.i.d. samples from an arbitrary distribution $g$ for some $\epsilon,\delta>0$. Then, there exist a deterministic algorithm $\mathscr{A}$, which outputs a number $j \in \{1,\ldots,M\}$ satisfying the following inequality with probability at least $1-\delta$:
\begin{equation}
\| f_j - g \|_{TV} \leq 3\cdot \min_{i \in \{1,2,\cdots,M\}} \|f_i - g\|_{TV} +4\epsilon.
\end{equation}
\end{theorem}
Proof can be found inside the reference. Combining the results of Theorem \ref{combinatoeic algorithm} and Lemmas \ref{lemma2} and \ref{quantization lemma}, we can present one of our main results as follows: The set of all shape-restricted simplices in $\mathbb{R}^K$ which entirely fall inside a sphere with radius $R$ is PAC-learnable:

\begin{theorem}[PAC-Learnability of Simplices in $\mathrm{C}^{K}(\boldsymbol{p}, R)$ Corrupted by Gaussaim Noise with Bounded Variance]
\label{thm:PAC:mainFinalPAC}
Assume $\boldsymbol{p}\in\mathbb{R}^K$, and $R,R_n>0$. Then, the class of $\left(\underline{\theta}, \bar{\theta}\right)$-isoperimetric $K$-simplices contained in the $K$-dimensional hyper-sphere $\mathrm{C}^{K}(\boldsymbol{p}, R)$ and convolved with an isotropic Gaussian noise $\mathcal{N}\left(\boldsymbol{0}, \sigma^2 \boldsymbol{\mathrm{I}}\right)$, with $\sigma \leq \mathrm{R}_n$, is PAC-learnable. Specifically, for some $\epsilon,\delta>0$, assume we have at least $n$ i.i.d. samples from a distribution $f_{\mathcal{S}}*G_{\sigma}$ with $\sigma \leq \mathrm{R}_n$ and $\mathcal{S} \in \mathrm{C}^{K}(\boldsymbol{p}, R)$, where the the following bound is satisfied:
\begin{equation}
n \geq 50\frac{\log{\frac{30R_n\sqrt{K}}{\delta\epsilon}} + 2(K+1)^2\log \left(1+ \frac{100R\bar{\theta}(K+1)}{\epsilon\mathrm{Vol}\left(\mathcal{S}\right)^{\frac{1}{K}}}\right)}{\epsilon^2}.
\end{equation}
Then, there exists an algorithm $\mathscr{A}$ whose outputs $\mathcal{S}_{\mathscr{A}}$ and $\sigma_{\mathscr{A}}$ satisfy the following inequality with probability at least $1-\delta$:
\begin{equation}
\label{corrolary1}
\left\Vert
f_{\mathcal{S}}*G_{\sigma}
-
f_{\mathcal{S}_{\mathscr{A}}}*G_{\sigma_{\mathscr{A}}}
\right\Vert_{\mathrm{TV}} \leq \epsilon.
\end{equation}
\end{theorem}
The proof can be found in Section \ref{sec:proof of theorems} of the Appendix.
Let $\mathbb{G}_{K,\sigma}\left(\boldsymbol{p},R\right)$ represent the distribution set
$$
\left\{f_{\mathcal{S}}*G_{\sigma}
\vert
\mathcal{S}\subseteq C^K\left(\boldsymbol{p},R\right)
\right\}.
$$
In an agnostic setting, so far we have actually proved that for any $\epsilon,R,R_n>0$, $\boldsymbol{p}\in\mathbb{R}^K$, and having $n\ge\tilde{\Omega}\left(K^2\log R/\epsilon^2\right)$ samples from ``any" distribution in
$$
\bigcup_{\sigma\leq R_n}\mathbb{G}_{K,\sigma}\left(\boldsymbol{p},R\right),
$$
there exists an algorithm $\mathscr{A}$ that outputs a simplex $\mathcal{S}_{\mathscr{A}}$ and noise standard deviation $\sigma_{\mathscr{A}}$ such that with a positive probability we have
\begin{align}
&
\left\Vert
f_{\mathcal{S}_{\mathcal{A}}}*G_{K,\sigma_{\mathscr{A}}}-
f_{\mathcal{S}}*G_{\sigma}
\right\Vert_{\mathrm{TV}}
\nonumber\\
&\quad\leq
4\min_{
\mathcal{S}^*\subseteq\mathrm{C}^{K}(\boldsymbol{p}, R)
\atop
\sigma^*\leq R_n}
\left\Vert
f_{\mathcal{S}^*}*G_{\sigma^*}-f_{\mathcal{S}}*G_{\sigma}
\right\Vert_{\mathrm{TV}}
+
\epsilon,
\end{align}
while the first term in the r.h.s. becomes zero in the realizable case. Also, note that $\bar{\theta}$ which represents the regularity level in the shape of simplices is also present inside the bounds. In fact, trying to learn a highly stretched simplex from noisy data can become very tricky since even a small amount of noise can shoot almost all the samples outside of the simplex. 

Another limitation of Theorem \ref{corrolary1} is that it requires the simplices to be inside an sphere of radius $R$. Using Lemma \ref{lemma2}, we show that this is not an actual necessity. The following theorem completes our first main result in this paper, which is the complete PAC-learnability of the class of noisy simplices:
\begin{theorem}[PAC-Learnability of Noisy Simplices in General]
\label{Theorem2}
The class of $\left(\underline{\theta}, \bar{\theta}\right)$-isoperimetric $K$-simplices which are corrupted with Gaussian white noise, i.e.,
$$
\bigcup_{\sigma>0}\mathbb{G}_{K,\sigma}
$$
is PAC-learnable. In other words, assume $n\geq \tilde{O}\left({K^2}/{\varepsilon^2}\right)$ i.i.d. samples from a noisy simplex $f_{\mathcal{S}}*G_{\sigma}$, for any $\mathcal{S}\in\mathbb{S}_K$ and $\sigma>0$, are given. Then, there exists an algorithm $\mathscr{A}$ which outputs a noisy simplex $f_{\mathcal{S}_{\mathscr{A}}}*G_{\sigma_{\mathscr{A}}}$ which with high probability satisfies
\begin{equation}
\|
f_{\mathcal{S}_{\mathscr{A}}}*G_{\sigma_{\mathscr{A}}}
-
f_{\mathcal{S}}*G_{\sigma}
\|_{\mathrm{TV}}
\leq \varepsilon.
\end{equation}
\end{theorem}
Proof of the above theorem can be found in Section \ref{sec:proof of theorems} of the Appendix. Obviously, the $\log R$ dependence inside the sample complexity of Theorem \ref{corrolary1} has been disappeared in that of Theorem \ref{Theorem2}.

\subsection{Denoising}

So far, we have shown that for a simplex $\mathcal{S}\in\mathbb{S}_K$ with some levels of geometric regularity, one can learn $f_{\mathcal{S}}*G_{\sigma}$ up to an arbitrarily small error as long as $n$ satisfies the bound in Theorem \ref{thm:PAC:mainFinalPAC}. What remains to prove is that a consistent estimation of $f_{\mathcal{S}}*G_{\sigma}$ leads to the same type of estimate for $f_{\mathcal{S}}$ as well. Our approach is based on showing the following three important properties:
\begin{itemize}
\item 
The difference function between any two distinct and geometrically regular members of
$\left\{f_{\mathcal{S}}\vert~\mathcal{S}\in\mathbb{S}_K\right\}$
corresponds to a {\it {low-frequency}} function, where ``low-frequency" refers to the Fourier domain. In other words, it means that the majority of the energy of the difference function lies in those regions of the Fourier domain which are close to the origin.

\item
Being corrupted by an additive noise is equivalent to convolving each $f_{\mathcal{S}}$ with $G_{\sigma}$. Moreover, convolution transforms into point-wise multiplication in the Fourier domain.

\item 
Finally, convolution with a Gaussian kernel $G_{\sigma}$ preserves the low-frequency parts of the difference function. Therefore, the two simplices remain distinguishable even after corruption via additive Gaussian noise.
\end{itemize}
Mathematically speaking, assume $\mathcal{S}_1,\mathcal{S}_2\in\mathbb{S}_K$ are distinct and have a minimum degree of geometric regularity, e.g., $\mathcal{L}_{\max}\left(\mathcal{S}_i\right)$ is bounded for $i=1,2$. Our aim is to show that if $f_{\mathcal{S}_1}*G_{\sigma}$ and $f_{\mathcal{S}_2}*G_{\sigma}$ have a total variation distance of at least $\epsilon>0$, then the $\ell_2$-distance between $f_{\mathcal{S}_1}$ and $f_{\mathcal{S}_2}$ is also bounded away from zero according to a function of $\epsilon,\sigma$ and the geometric regularity of simplices $\mathcal{S}_1$ and $\mathcal{S}_2$. The theoretical core behind our method is stated in the following general theorem.

\begin{theorem}[Recovery of Low-Frequency Objects from Additive Noise]
\label{thm:generalResultTheorem}
For $K\in\mathbb{N}$, consider a probability density function family $\mathscr{F}$ which is supported over $\mathbb{R}^K$. Assume for sufficiently large $\alpha>0$, the following bound holds for all $f,g\in\mathscr{F}$:
\begin{align*}
&\frac{1}{\left(2\pi\right)^K}
\int_{\left\Vert\boldsymbol{\omega}\right\Vert_{\infty}\ge\alpha}
\left\vert
\mathcal{F}\left\{f-g\right\}\left(\boldsymbol{\omega}\right)
\right\vert^2
\leq
\zeta\left(\alpha^{-1}\right)
\int_{\mathbb{R}^K}
\left\vert
f-g
\right\vert^2,
\end{align*}
where $\mathcal{F}\left\{\cdot\right\}$ denotes the Fourier transform and $\zeta$ is an increasing function with $\zeta\left(0\right)=0$ and continuity at $0$. Also, assume the probability density function $Q$ 
(also supported over $\mathbb{R}^K$) has the following property:
\begin{equation*}
\inf_{\left\Vert\boldsymbol{\omega}\right\Vert_{\infty}\leq\alpha}
\left\vert
\mathcal{F}\left\{Q\right\}\left(\boldsymbol{\omega}\right)
\right\vert
\ge \eta\left(\alpha\right),
\end{equation*}
where $\eta\left(\cdot\right)$ is a non-negative decreasing function. Then, there exists a non-negative constant $C$ where for any $\sigma,\varepsilon>0$ and $f,g\in\mathscr{F}$ with
$
\left\Vert
f-g
\right\Vert_2
\geq\varepsilon,
$
we have
$$
\left\Vert
\left(f-g\right)*Q
\right\Vert_2
\geq
\frac{\varepsilon}{\left(2\pi\right)^{K/2}}
\left(
\sup_{\alpha\ge0}~
\eta\left(\alpha\right)
\sqrt{1-\zeta\left(\alpha^{-1}\right)}
\right),
$$
\end{theorem}
\begin{proof}
For the sake of simplicity in notations, let $\mathcal{F},\mathcal{G},\mathcal{Q}:\mathbb{R}^K\rightarrow\mathbb{C}$ denote the Fourier transforms of $f,g$ and $Q$, respectively. Due to Parseval's theorem, we have
$$
\left\Vert
f-g
\right\Vert^2_2
=
\frac{1}{\left(2\pi\right)^K}
\left\Vert
\mathcal{F}-\mathcal{G}
\right\Vert^2_2.
$$
Also, due to the properties of the Fourier transform, which is the transformation of convolution into direct multiplication, one can write
$$
\mathcal{F}\left\{
\left(f-g\right)*Q
\right\}
=
\mathcal{Q}\left(\mathcal{F}-\mathcal{G}\right).
$$
Thus, for sufficiently large $\alpha$ we have:
\begin{align}
\left(2\pi\right)^K
&
\left\Vert
\left(f-g\right)*Q
\right\Vert^2_2
=
\int_{\mathbb{R}^K}
\left\vert
\mathcal{Q}
\left(\boldsymbol{\omega}\right)
\left(
\mathcal{F}\left(\boldsymbol{\omega}\right)
-
\mathcal{G}\left(\boldsymbol{\omega}\right)
\right)
\right\vert^2
\nonumber\\
&\quad\quad\quad\ge
\int_{\left\Vert\boldsymbol{\omega}\right\Vert_{\infty}\leq\alpha}
\left\vert
\mathcal{Q}
\left(\boldsymbol{\omega}\right)
\right\vert^2
\left\vert
\mathcal{F}\left(\boldsymbol{\omega}\right)
-
\mathcal{G}\left(\boldsymbol{\omega}\right)
\right\vert^2
\nonumber\\
&\quad\quad\quad\ge
\eta^2\left(\alpha\right)
\int_{\left\Vert\boldsymbol{\omega}\right\Vert_{\infty}\leq\alpha}
\left\vert
\mathcal{F}\left(\boldsymbol{\omega}\right)
-
\mathcal{G}\left(\boldsymbol{\omega}\right)
\right\vert^2
\nonumber\\
&\quad\quad\quad\geq
\varepsilon^2
\eta^2\left(\alpha\right)
\left[
1-\zeta\left(\alpha^{-1}\right)
\right].
\end{align}
The above chain of inequalities hold for all $\alpha>C$, therefore we have:
\begin{align*}
\left\Vert
\left(f-g\right)*Q
\right\Vert_2
&\ge
\frac{\varepsilon}{\left(2\pi\right)^{K/2}}
\sup_{\alpha}~
\eta\left(\alpha\right)
\sqrt{1-\zeta\left(\alpha^{-1}\right)},
\end{align*}
which completes the proof.
\end{proof}


Theorem \ref{thm:generalResultTheorem} presents a general approach to prove the recoverability of latent functions (or objects, which are the main focus in this work) from a certain class of independent additive noise distributions. This approach works as long as the function class as well as the noise distribution are mostly comprised of low-frequency components in the Fourier domain. For example, the Gaussian noise hurts low-frequency parts of a geometric object far less than its high-frequency details. More specifically, we prove the following corollary for Theorem \ref{thm:generalResultTheorem}:
\begin{corollary}[Recoverability from Additive Gaussian Noise $\mathcal{N}\left(\boldsymbol{0},\sigma^2\boldsymbol{I}\right)$]
\label{corl:GaussinNoiseMain}
Consider the setting in Theorem \ref{thm:generalResultTheorem}, and assume the noise distribution follows $Q\triangleq\mathcal{N}\left(\boldsymbol{0},\sigma^2\boldsymbol{I}\right)$ for $\sigma>0$. Then, as long as for $f,g\in\mathscr{F}$ we have $\left\Vert f-g\right\Vert_2\ge\varepsilon$ for some $\varepsilon\ge0$, we also have
\begin{align}
&\left\Vert
\left(f-g\right)*Q
\right\Vert_2
\ge
\nonumber\\
&\quad\quad
\frac{\varepsilon}{\left(2\pi\right)^{K/2}}
\left(
\sup_{\alpha>C}~
\sqrt{1-\zeta\left(\frac{1}{\alpha}\right)}
e^{-K\left(\sigma\alpha\right)^2/2}
\right)
\end{align}
\end{corollary}
\begin{proof}
The Fourier transform of $Q=\mathcal{N}\left(\boldsymbol{0},\sigma^2\boldsymbol{I}\right)$ can be computed as follows:
\begin{equation*}
\mathcal{F}\left\{Q\right\}\left(\boldsymbol{\omega}\right)
=
\prod_{i=1}^{K}\mathcal{F}\left\{\mathcal{N}\left(0,\sigma^2\right)\right\}\left(\omega_i\right)
=
e^{-\sigma^2\left\Vert\boldsymbol{\omega}\right\Vert_2^2/2}.
\end{equation*}
Also, it can be easily checked that
$$
\inf_{\left\Vert\boldsymbol{\omega}\right\Vert_{\infty}\leq\alpha}
e^{-\sigma^2\left\Vert\boldsymbol{\omega}\right\Vert_2^2/2}=
e^{-\sigma^2/2\left(\alpha^2+\ldots+\alpha^2\right)}
=
e^{-K\left(\alpha\sigma\right)^2/2}.
$$
By subsititution into the end result of Theorem \ref{thm:generalResultTheorem}, the claimed bounds can be achieved and the proof is complete.
\end{proof}


In this regard, our main explicit theoretical contribution in this section with respect to simplices has been stated in the following theorem:

\begin{theorem}[Recoverability of Simplices from Additive Noise]
\label{thm:mainNoisySimplexBound}
For any two $\left(\bar{\theta},\underline{\theta}\right)$-isoperimetric simplices $\mathcal{S}_1,\mathcal{S}_2\in\mathbb{S}_K$, given that
$$
\|
f_{\mathcal{S}_1}*G_{\sigma}-f_{\mathcal{S}_2}*G_{\sigma}
\|_{\mathrm{TV}}
\leq
\varepsilon
$$
for some $\varepsilon\ge0$, we have
$$
\|
f_{\mathcal{S}_1}-f_{\mathcal{S}_2}\|_2
\leq \varepsilon
e^{\Omega\left(\frac{K}{\mathrm{SNR}^2}\right)},
$$
where $\mathrm{SNR}$ denotes the effective signal-to-noise ratio which is defined earlier.
\end{theorem}
Proof is given in Section C of the supplementary document. The core idea is to first showing that simplices, in general, are low-frequency objects in the Fourier domain which holds due to their convexity and simple geometric shape. More specifically, Lemma C.3 shows that for a geometrically regular simplex $\mathcal{S}\in\mathbb{S}_K$, we have
$$
\frac{1}{\left(2\pi\right)^K}
\int_{\left\Vert\boldsymbol{\omega}\right\Vert_{\infty}\ge\alpha}
\left\vert
\mathcal{F}\left\{f_{\mathcal{S}}\right\}\left(\boldsymbol{\omega}\right)
\right\vert^2
\leq
\frac{1}{\mathrm{Vol}\left(\mathcal{S}\right)}
\mathcal{O}\left(\frac{K}{\alpha}\right),
$$
for a sufficiently large $\alpha>0$, where constants in $\mathcal{O}\left(\cdot\right)$ only depend on regularity parameters of $\mathcal{S}$. This inequality is trivial when $K=1$, since a one-dimensional simplex is a pulse function whose Fourier transform is a $\mathrm{sinc}(x)=\frac{\sin ax}{ax}$ for some $a>0$. However, showing this relation for larger values of $K$ requires more mathematical effort which is already carried out in the proof of Theorem \ref{thm:mainNoisySimplexBound}. Furthermore, we show a similar inequality holds for the difference of simplices as well. This would ultimately enable us to use the result of Corollary \ref{corl:GaussinNoiseMain} to prove the main claim of the paper.


\section{Conclusions}
\label{sec:conc}
We present the first sample complexity bounds for consistent learning of high-dimensional simplices in noisy regimes. Formally speaking, we prove that given a sufficient amount of noisy data, one can estimate the true simplex up to an arbitrarily small $\ell_2$-error. Also, a presumably optimal polynomial dependence on a number of parameters, such as dimension has been achieved which matches those of the already-solved noiseless case. More interestingly, we have found theoretical justification for an already observed phenomenon in practice: The performance of most heuristic methods in learning of high-dimensional simplices undergo a rather sharp phase transition w.r.t. the noise power, where for $\mathrm{SNR}\ge\Omega\left(K^{1/2}\right)$ they behave almost identical to the noiseless case, but their performance severely degrades when $\mathrm{SNR}$ becomes smaller. Our proofs are based on a number of recent techniques which have been previously used on learning of Gaussian mixture models, a number of tools from high-dimensional geometry and a novel Fourier-based technique for denoising. The latter technique has applications a broader spectra of problems where general recovery of distribution families with certain low-frequency properties from additive Gaussian noise is under question. For future directions in this area, one can think of analyzing this problem under a broader noise and distortion model which might match the practice even more. Also, finding the first non-trivial and amenable polynomial-time algorithms for this problem is an interesting line of work which has remained open until this day.

\section{Acknowledgements}

This research was partially supported by a grant from IPM.


\bibliography{ref}
\bibliographystyle{icml2023}

\newpage
\appendix
\onecolumn

\section{Proof of Theorems}
\label{sec:proof of theorems}
\begin{proof}[proof of Theorem \ref{thm:PAC:mainFinalPAC}]
Suppose that $\mathcal{S}$ is a $\left(\underline{\theta}, \bar{\theta}\right)$-isoperimetric simplex in $\mathbb{R}^K$ which is confined in a $K$-dimensional hyper-sphere $\mathrm{C}^{K}(\boldsymbol{p}, R)$. From Lemma \ref{quantization lemma}, we know how to build an $\epsilon$-representative set, for any $\epsilon>0$, for all $\left(\underline{\theta}, \bar{\theta}\right)$-isoperimetric $K$-simplices in $\mathrm{C}^{K}(\boldsymbol{p}, R)$. Let us denote this set with $\widehat{\mathbb{S}}(\mathrm{C}^{K}(\boldsymbol{p}, R))$. 

Now Consider the class of all isotropic Gaussian noise, $\mathcal{N}\left(\boldsymbol{0},\sigma^2\boldsymbol{I} \right)$, with $\sigma \leq R_n$. We denote this class of distributions with $\mathfrak{N}^K\left(R_n\right)$. Consider a set $C^{\epsilon/\sqrt{K}}_{R_n} = \left\{0, \frac{\epsilon}{\sqrt{K}}, \frac{2\epsilon}{\sqrt{K}}, \cdots, R_n\right\}$, which $\frac{\epsilon}{\sqrt{K}}$-covers the interval $\left[0, R_n\right]$. Now for all $\sigma_i \in C^{\epsilon/\sqrt{K}}_{R_n}$ we put $\mathcal{N}\left(\boldsymbol{0},\sigma_i^2\boldsymbol{I} \right)$ in a set called $\widehat{\mathfrak{N}}^K\left(R_n\right)$. From Theorem $1.1$ in \citep{devroye2018total} it can be shown that $\widehat{\mathfrak{N}}^K\left(R_n\right)$ is an $\epsilon$-representative set for $\mathfrak{N}^K\left(R_n\right)$. According to the way we build $\widehat{\mathfrak{N}}^K\left(R_n\right)$, we have $\vert\widehat{\mathfrak{N}}^K\left(R_n\right)\vert \leq \frac{R_n\sqrt{K}}{\epsilon}$. Now we build a set of noisy simplices as follows:
\begin{equation}
    \widehat{\mathbb{G}}_{K, \sigma}(R, R_n) \triangleq \left\{f_{\mathcal{S}}\left(\boldsymbol{x}\right) \ast G_{\sigma}\left(\boldsymbol{x}\right)\vert \mathcal{S} \in \widehat{\mathbb{S}}(\mathrm{C}^{K}(\boldsymbol{p}, R)), G_{\sigma} \in \widehat{\mathfrak{N}}^K\left(R_n\right)\right\}.
\end{equation}
Now for any density function $f_{\mathcal{S}}\left(\boldsymbol{x}\right) \ast G_{\sigma}\left(\boldsymbol{x}\right)$, where $\mathcal{S} \in \mathrm{C}^{K}(\boldsymbol{p}, R)$ and $\sigma \leq R_n$, we can find some $\mathcal{S}^{\star} \in \widehat{\mathbb{S}}(\mathrm{C}^{K}(\boldsymbol{p}, R))$, and $G_{\sigma^\star} \in \widehat{\mathfrak{N}}^K\left(R_n\right)$ such that
\begin{align*}
    &\|f_{\mathcal{S}}- f_{\mathcal{S}^{\star}}\|_{\mathrm{TV}} \leq \epsilon,
    \\
    &\|G_{\sigma}- G_{\sigma^{\star}}\|_{\mathrm{TV}} \leq \epsilon,
    \\
    &f_{\mathcal{S}^{\star}} \ast G_{\sigma^{\star}} \in \widehat{\mathbb{G}}_{K, \sigma}(R, R_n).
\end{align*}
Now for the distance between $f_{\mathcal{S}^{\star}}^{\sigma^{\star}}$ and $f_{\mathcal{S}}^{\sigma}$ we have:
\begin{align}
    \|f_{\mathcal{S}^{\star}} \ast G_{\sigma^{\star}}-f_{\mathcal{S}} \ast G_{\sigma}\|_{\mathrm{TV}}
    \leq & \|f_{\mathcal{S}^{\star}} \ast G_{\sigma^{\star}}- f_{\mathcal{S}^{\star}} \ast G_{\sigma}\|_{\mathrm{TV}} + \|f_{\mathcal{S}^{\star}} \ast G_{\sigma}- f_{\mathcal{S}} \ast G_{\sigma}\|_{\mathrm{TV}}
    \nonumber \\
    \leq & \|G_{\sigma^{\star}} - G_{\sigma}\|_{\mathrm{TV}} + \|f_{\mathcal{S}^{\star}} - f_{\mathcal{S}} \|_{\mathrm{TV}}
    \nonumber \\ 
    \leq & 2\epsilon.
\end{align}
From the above inequalities it can be seen that for any density function $f_{\mathcal{S}} \ast G_{\sigma}$, where $\mathcal{S} \in \mathrm{C}^{K}(\boldsymbol{p}, R)$ and $\sigma \leq R_n$, there exist some density function $f^{\star} \in \widehat{\mathbb{S}}_n(\mathrm{C}^{K}(\boldsymbol{p}, R), R_n)$
where $\|f^{\star} - f_{\mathcal{S}} \ast G_{\sigma}\|_{\mathrm{TV}} \leq 2\epsilon$. Therefore the set $ \widehat{\mathbb{G}}_{K, \sigma}(R, R_n)$ is a $2\epsilon$-representative set for the class of $K$-simplices confined in a hyper-sphere with radius $R$ which convolved with a Gaussian noise with variance $\sigma \leq R_n$. We show this class with 

\begin{equation}
    \mathbb{G}_{K,\sigma}\left(R, R_n\right)
\triangleq
\left\{
f_{\mathcal{S}}*G_{\sigma}
\vert~
\mathcal{S}\in\mathbb{S}_K, \mathcal{S} \in \mathrm{C}^{K}(\boldsymbol{p}, R), \sigma \leq R_n
\right\}
\end{equation}

Assume that we have a set of i.i.d. samples from some distribution $f_{\mathcal{S}}*G_{\sigma} \in \mathbb{G}_{K,\sigma}\left(R, R_n\right)$. From Theorem \ref{combinatoeic algorithm}, we know that there exists a deterministic algorithm $\mathscr{A}$ such that given
$$n \geq \frac{\log{(3\vert\widehat{\mathbb{G}}_{K, \sigma}(R, R_n)\vert^2/\delta)}}{2\epsilon^2}$$
i.i.d. samples from $f_{\mathcal{S}}*G_{\sigma}$, the output of the algorithm denoted by  $f_{\mathcal{S}_{\mathscr{A}}}*G_{\sigma_{\mathscr{A}}}$, with probability at least $1-\delta$, satisfies:
\begin{align}
\|f_{\mathcal{S}_{\mathscr{A}}}*G_{\sigma_{\mathscr{A}}} -  f_{\mathcal{S}}*G_{\sigma} \|_{\mathrm{TV}}  \leq &~
3 \min_{f \in \widehat{\mathbb{G}}_{K, \sigma}(R, R_n)} \|f - f_{\mathcal{S}}*G_{\sigma}\|_{\mathrm{TV}} +4\epsilon
\nonumber \\
 \leq&~ 6\epsilon +4\epsilon = 10\epsilon.
 \label{combinatorial method 2}
\end{align}
For the cardinality of $\widehat{\mathbb{G}}_{K, \sigma}(R, R_n)$ we have:
\begin{align}
    \vert\widehat{\mathbb{G}}_{K, \sigma}(R, R_n)\vert 
    =& \vert\widehat{\mathbb{S}}(\mathrm{C}^{K}(\boldsymbol{p}, R))\vert \vert\widehat{\mathfrak{N}}^K\left(R_n\right)\vert 
    \nonumber \\
    \leq& \vert\widehat{\mathbb{S}}(\mathrm{C}^{K}(\boldsymbol{p}, R))\vert \frac{R_n\sqrt{K}}{\epsilon}.
    \label{noisy representative cardinality}
\end{align}
And, according to Lemma \ref{quantization lemma} the followings hold for the cardinality of $\widehat{\mathbb{S}}(\mathrm{C}^{K}(\boldsymbol{p}, R))$: 
\begin{align}
\bigg\vert\widehat{\mathbb{S}}(\mathrm{C}^{K}(\boldsymbol{p}, R))\bigg\vert = &
\nonumber
 \binom{\left\vert \mathrm{T}_{\frac{\alpha\epsilon}{K+1}}(\mathrm{C}^{K}(\boldsymbol{p}, R))\right\vert}{K+1} 
 \\ \nonumber \leq &~ \bigg\vert \mathrm{T}_{\frac{\alpha\epsilon}{K+1}}(\mathrm{C}^{K}(\boldsymbol{p}, R))\bigg\vert^{K+1}
 \\ \nonumber \leq &~ \left(\left(1+ \frac{2(K+1) R}{\alpha\epsilon} \right)^K\right)^{K+1}
  \\   = &~ \left(1+ \frac{2(K+1)R}{\alpha\epsilon} \right)^{K(K+1)}.
  \label{quantized simplices set cardinality}
\end{align}
Now from \ref{quantized simplices set cardinality} and \ref{noisy representative cardinality} we have:
\begin{align}
    \vert\widehat{\mathbb{G}}_{K, \sigma}(R, R_n)\vert
    \leq & \frac{R_n\sqrt{K}}{\epsilon} \left(1+ \frac{2(K+1)R}{\alpha\epsilon} \right)^{K(K+1)}.
    \label{quantized noisy simplices cardinality}
\end{align}
Then using \ref{quantized noisy simplices cardinality} and \ref{combinatorial method 2}, we can say that for any $\epsilon_,\delta>0$, there exists a PAC-learning algorithm $\mathscr{A}$ for the class of noisy $\left(\underline{\theta}, \bar{\theta}\right)$-isoperimetric $K$-simplices in $\mathbb{G}_{K,\sigma}\left(R, R_n\right)$, whose sample complexity is bounded as follows:
\begin{align}
n \geq &~ 50\frac{\log{\frac{10R_n\sqrt{K}}{\epsilon}} + 2(K+1)^2\log \left(1+ \frac{20(K+1)R}{\alpha\epsilon} \right) + \log\frac{3}{\delta}}{\epsilon^2}
\nonumber \\
= &~ 50\frac{\log{\frac{30R_n\sqrt{K}}{\delta\epsilon}} + 2(K+1)^2\log \left(1+ \frac{100R\bar{\theta}(K+1)}{\epsilon\mathrm{Vol}\left(\mathcal{S}\right)^{\frac{1}{K}}}\right)}{\epsilon^2}.
\end{align}
In other words, given that the number of samples $n$ satisfies the above lower-bound, then with probability at least $1-\delta$ we have
$$
\|f_{\mathcal{S}_{\mathscr{A}}}*G_{\sigma_{\mathscr{A}}} - f_{\mathcal{S}}*G_{\sigma} \|_{\mathrm{TV}} \leq \epsilon.
$$
This completes the proof.
Using simple algebra, the bound can be further simplified into $n\ge O\left(\frac{K^2}{\epsilon^2}\log{\frac{K}{\epsilon}}\right)$ which completes the proof.
\end{proof}

\begin{proof}[proof of Theorem \ref{Theorem2}]
From Theorem \ref{thm:PAC:mainFinalPAC}, we know that the class of $\left(\underline{\theta}, \bar{\theta}\right)$-isoperimetric $K$-simplices contained in the $K$-dimensional hyper-sphere $\mathrm{C}^{K}(\boldsymbol{p}, R)$ and convolved with an isotropic Gaussian noise $\mathcal{N}\left(\boldsymbol{0}, \sigma^2 \boldsymbol{\mathrm{I}}\right)$, with $\sigma \leq \mathrm{R}_n$, is PAC-learnable, with sample complexity $O\left(\frac{K^2}{\epsilon^2}\log{\frac{K}{\epsilon}}\right)$, where $\epsilon$ is defined accordingly. And from Lemma \ref{lemma2}, we know that if we have $O\left(K^2\right)$ samples from a noisy simplex $f_{\mathcal{S}}*\mathbb{G}_\sigma$, we can find a $K$-dimensional sphere which with probability at least $1-\delta/2$ contains the true simplex as long as the radius $R$ of the sphere and the upper-bound of the noise variance $R_n$ satisfy
\begin{align}
& R \leq 4\sqrt{K+1}\left(1+\frac{K+2}{d_{\mathrm{S}}/\sigma}\right)d_{\mathrm{S}}
\label{upper bound for the radius of containing sphere}
\\
& R_n \leq \frac{K+2}{K-3}\left(1+\frac{d_{\mathrm{S}}/\sigma}{K+2}\right)\sigma.
\label{upper bound for the noise variance}
\end{align}

Therefore, it can be shown that there exists an algorithm $\mathscr{A}$ such that given $n$ i.i.d samples from $\mathbb{G}_\mathcal{S}$ with
\begin{align}
n
\geq&~ 50\frac{\log{\frac{30R_n\sqrt{K}}{\delta\epsilon}} + 2(K+1)^2\log \left(1+ \frac{100R\bar{\theta}(K+1)}{\epsilon\mathrm{Vol}\left(\mathcal{S}\right)^{\frac{1}{K}}}\right)}{\epsilon^2}
+ 2000(K+1)(K+2)\log{\frac{6}{\delta}}
\nonumber\\
=&~ 100\frac{\log{6/\delta}+(K+2)^2\log \left(1+ \frac{100\bar{\theta}(K+1)^{3/2}}{\epsilon\mathrm{Vol}\left(\mathcal{S}\right)^{\frac{1}{K}}}\left(1+\frac{K+2}{d_{\mathrm{S}}/\sigma}\right)d_{\mathrm{S}}\right)}{\epsilon^2}
\nonumber\\
=&~ O\left(\frac{K^2}{\epsilon^2}\log\frac{K}{\epsilon}\right),
\label{sample complexity of noisy simplex}
\end{align}
the output of the algorithm $f_{\mathcal{S}_{\mathscr{A}}}*G_{\sigma_{\mathscr{A}}}$ with probability at least $1-\delta$ satisfies
\begin{equation*}
\|f_{\mathcal{S}_{\mathscr{A}}}*G_{\sigma_{\mathscr{A}}} - f_{\mathcal{S}}*\mathbb{G}_\sigma\|_\mathrm{TV} \leq ~\epsilon.
\end{equation*}
It should be noted that $R$ and $R_n$ in \eqref{sample complexity of noisy simplex} is already replaced with the bound in the r.h.s of \eqref{upper bound for the radius of containing sphere} and \eqref{upper bound for the noise variance}.
This way, the proof is completed.
\end{proof}
\section{Proof of Lemmas}
\label{sec:proof of lemmas}
\begin{proof}[proof of Lemma \ref{lemma2}]
To find a $K$-dimensional sphere containing the true simplex, it suffices to find a point $\boldsymbol{p}$ inside the simplex and an upper-bound $R$ for its diameter, i.e., the maximum distance between two points in the simplex. Obviously, this ensures that the $K$-dimensional sphere with center point $\boldsymbol{p}$ and radius $R$ contains the true simplex. For the maximum distance between any two points inside $\mathcal{S}\in\mathbb{S}_K$, denoted by $d_{\mathcal{S}}$, we have
\begin{align}
d_{\mathcal{S}} = &\max_{\boldsymbol{x},\boldsymbol{y} \in \mathcal{S}}{\|\boldsymbol{x}-\boldsymbol{y}\|_2} 
\nonumber \\ 
=& \max_{\boldsymbol{\phi}_{\boldsymbol{x}}, \boldsymbol{\phi}_{\boldsymbol{y}} \in \mathcal{S}_\mathrm{S}^K}{\|\boldsymbol{V}_{\mathcal{S}}\boldsymbol{\phi}_{\boldsymbol{x}} - \boldsymbol{V}_{\mathcal{S}}\boldsymbol{\phi}_{\boldsymbol{y}} \|_2}
\nonumber \\ 
=& \max_{\boldsymbol{\phi}_{\boldsymbol{x}}, \boldsymbol{\phi}_{\boldsymbol{y}} \in \mathcal{S}_\mathrm{S}^K}{\|\boldsymbol{V}_{\mathcal{S}}\left(\boldsymbol{\phi}_{\boldsymbol{x}} -\boldsymbol{\phi}_{\boldsymbol{y}}\right) \|_2},
\end{align}
where $\mathcal{S}_{\mathrm{S}}^K$ represents the standard simplex in $\mathbb{R}^K$ and $\boldsymbol{V}_{\mathcal{S}}$ denotes the vertex matrix of $\mathcal{S}$ (Equation 4 in the main paper). Let $\sigma_{\max}\left(\mathcal{S}\right)$ denote the maximum singular-value for the matrix $\boldsymbol{V}_{\mathcal{S}}$, and assume $\boldsymbol{v}_{\max}$ be the unitary eigenvector that corresponds to $\sigma_{\max}\left(\mathcal{S}\right)$. To be more precise, let $\boldsymbol{V}_{\mathcal{S}}=\boldsymbol{U}\boldsymbol{\Sigma}\boldsymbol{V}^T$ be the singular value decomposition of the vertex matrix. Then, $\sigma_{\max}\left(\mathcal{S}\right)$ refers to the largest singular value on the main diagonal of $\boldsymbol{\Sigma}$, and $\boldsymbol{v}_{\max}$ represents the corresponding column in $\boldsymbol{V}$. Then, for any point $\boldsymbol{x}\in\mathbb{R}^{K+1}$, one can write
$$
\left\Vert
\boldsymbol{V}_{\mathcal{S}}\boldsymbol{x}
\right\Vert^2_2
=\sum_{i}\Sigma^2_{i,i}\left\Vert
\boldsymbol{U}_i\boldsymbol{V}_i^T\boldsymbol{x}
\right\Vert^2_2
=\sum_{i}\Sigma^2_{i,i}\left(\boldsymbol{V}_i^T\boldsymbol{x}\right)^2,
$$
which results in the following useful inequality:
$$
\left\Vert
\boldsymbol{V}_{\mathcal{S}}\boldsymbol{x}
\right\Vert_2
\ge
\sigma_{\max}\left(\mathcal{S}\right)
\left\vert
\boldsymbol{v}_{\max}^T\boldsymbol{x}
\right\vert.
$$
For $\boldsymbol{\phi}_{\boldsymbol{x}}\neq\boldsymbol{\phi}_{\boldsymbol{y}}$, let $\boldsymbol{r}\triangleq\left(\boldsymbol{\phi}_{\boldsymbol{x}}-\boldsymbol{\phi}_{\boldsymbol{y}}\right)/\left\Vert\boldsymbol{\phi}_{\boldsymbol{x}}-\boldsymbol{\phi}_{\boldsymbol{y}}\right\Vert_2$, and $\ell\triangleq\left\Vert\boldsymbol{\phi}_{\boldsymbol{x}}-\boldsymbol{\phi}_{\boldsymbol{y}}\right\Vert_2$. In other words, let $\boldsymbol{r}$ represent the unitary direction vector for the difference, and $\ell$ to denote the corresponding length. In this regard, we have
\begin{align}
\max_{\boldsymbol{\phi}_{\boldsymbol{x}},\boldsymbol{\phi}_{\boldsymbol{y}}\in\mathcal{S}^K_\mathrm{S}}
\left\Vert
\boldsymbol{V}_{\mathcal{S}}\left(
\boldsymbol{\phi}_{\boldsymbol{x}}-\boldsymbol{\phi}_{\boldsymbol{y}}
\right)
\right\Vert_2
\ge~
\sigma_{\max}\left(\mathcal{S}\right)
\max_{\boldsymbol{r},\ell}~
\ell
\left\vert
\boldsymbol{v}_{\max}^T\boldsymbol{r}
\right\vert
\ge~
\sigma_{\max}\left(\mathcal{S}\right)
\max_{\ell\vert~\boldsymbol{r}=\boldsymbol{v}_{\max}}~
\ell.
\end{align}
Now we should find a lower-bound for maximum possible $\ell$ that can be achieved for some fixed direction $\boldsymbol{r}=\boldsymbol{v}_{\max}$. To this aim, assume the difference vector $\boldsymbol{\phi}_{\boldsymbol{x}}-\boldsymbol{\phi}_{\boldsymbol{y}}$ starts from any arbitrary vertex, i.e., $\boldsymbol{\phi}_{\boldsymbol{x}}$ is a one-hot vector with $K$ components equal to zero and the remaining one equal to $1$. In this regard, the minimum possible $\ell$ that can be achieved (minimum is taken w.r.t. direction of the difference vector) occurs when the difference vector becomes perpendicular to its front facet. Since all $\boldsymbol{\phi}$ vectors belong to the standard simplex, such difference vectors can be easily shown to be of the form $\boldsymbol{\Delta}$ with
$$
\Delta_i=1\quad\mathrm{and}\quad\Delta_{j\vert j\neq i}=-1/K,
$$
for any $i\in0,\ldots,K$. The $\ell_2$-norm of all such vectors equals to $\sqrt{1+1/K}$. Thus, we have
\begin{align}
d_{\mathcal{S}}\ge~\sqrt{\frac{K+1}{K}}\sigma_{\mathrm{max}}\left(S\right)
\ge~\sigma_{\mathrm{max}}\left(S\right).
\end{align}
Assume we have $2m$ i.i.d. samples drawn from a noisy simplex $f_{\mathcal{S}}*\mathbb{G}_{\sigma}$, denoted by $\left\{\boldsymbol{y}\right\}^{2m}_{i=1}$. This way, we have $\boldsymbol{y}_i=\boldsymbol{V}_{\mathcal{S}}\boldsymbol{\phi}_i+\boldsymbol{z}_i$ for $i=1,\ldots,2m$, where $\boldsymbol{\phi}_i$ represents the weight vector for the $i$th sample (drawn from a uniform Dirichlet distribution), while $\boldsymbol{z}_i\sim\mathcal{N}\left(\boldsymbol{0},\sigma^2\boldsymbol{I}\right)$ is an independently drawn noise vector. Let us define $\mathrm{D}$ as
\begin{equation}
\mathrm{D}= \frac{1}{2m}\sum_{i =1}^{m}{\|\boldsymbol{y}_{2i} - \boldsymbol{y}_{2i-1}\|_2^2}.
\end{equation}
To find an upper-bound for $d_{\mathcal{S}}$, it is enough to find a high probability lower-bound for $\mathrm{D}$ in terms of $d_{\mathcal{S}}$. To this aim, we rewrite $\mathrm{D}$ as
\begin{align}
\mathrm{D} =~  & \frac{1}{2m}\sum_{i =1}^{m}{\|\boldsymbol{V}_{\mathcal{S}}\boldsymbol{\phi}_{2i} - \boldsymbol{V}_{\mathcal{S}}\boldsymbol{\phi}_{2i-1} + \boldsymbol{z}_{2i} - \boldsymbol{z}_{2i-1} \|_2^2}
\nonumber \\  
=~ & \frac{1}{2m}\sum_{i =1}^{m}{\|\boldsymbol{V}_{\mathcal{S}}\boldsymbol{\phi}_{2i} - \boldsymbol{V}_{\mathcal{S}}\boldsymbol{\phi}_{2i-1}\|_2^2} + \frac{1}{2m}\sum_{i=1}^{m}{\|\boldsymbol{z}_{2i} - \boldsymbol{z}_{2i-1} \|_2^2} 
\nonumber \\
& + \frac{1}{m}\sum_{i=1}^{m}{\left(\boldsymbol{V}_{\mathcal{S}}\boldsymbol{\phi}_{2i} - \boldsymbol{V}_{\mathcal{S}}\boldsymbol{\phi}_{2i-1}\right)^T\left(\boldsymbol{z}_{2i} - \boldsymbol{z}_{2i-1}\right)}
\nonumber \\  
= & ~ \frac{1}{2m}\sum_{i =1}^{m}{\|\boldsymbol{V}_{\mathcal{S}}\boldsymbol{\phi}_{2i} - \boldsymbol{V}_{\mathcal{S}}\boldsymbol{\phi}_{2i-1}\|_2^2} + \frac{2\sigma^2}{2m}\sum_{i=1}^{m}{\|\tilde{\boldsymbol{z}}_{2i} -  \tilde{\boldsymbol{z}}_{2i-1} \|_2^2}
\nonumber \\ 
& + \frac{\sigma\sqrt{2}}{m}\sum_{i=1}^{m}{\left(\boldsymbol{V}_{\mathcal{S}}\boldsymbol{\phi}_{2i} - \boldsymbol{V}_{\mathcal{S}}\boldsymbol{\phi}_{2i-1}\right)^T\left(\tilde{\boldsymbol{z}}_{2i} - \tilde{\boldsymbol{z}}_{2i-1}\right)}.
\label{Lower bound for statistic}
\end{align}
Let us denote the three terms in r.h.s. of \eqref{Lower bound for statistic} as $\mathrm{I}$, $\mathrm{II}$ and $\mathrm{III}$, respectively. Also, we have $\tilde{\boldsymbol{z}}_{i} = \boldsymbol{z}_i/\sigma$. To find a lower-bound for $\mathrm{D}$, we should find respective lower-bounds for $\mathrm{I}$, $\mathrm{II}$ and $\mathrm{III}$. First, let us discuss about $(\mathrm{I})$:
\begin{equation}
\mathrm{I} = \frac{1}{2m}\sum_{i =1}^{m}{\left\Vert
\boldsymbol{V}_{\mathcal{S}}\boldsymbol{\phi}_{2i} - \boldsymbol{V}_{\mathcal{S}}\boldsymbol{\phi}_{2i-1}
\right\Vert_2^2}  =  f_{\boldsymbol{V}_{\mathcal{S}}}\left(\boldsymbol{\phi}_1,\boldsymbol{\phi}_2,\cdots, \boldsymbol{\phi}_{2m}\right).
\end{equation}
For the term inside of the summation, we have
\begin{align}
\|\boldsymbol{V}_{\mathcal{S}}\boldsymbol{\phi}_{2i} - \boldsymbol{V}_{\mathcal{S}}\boldsymbol{\phi}_{2i-1}\|_2^2 
~\stackrel{a.s.}{\leq}~ & d_{\mathcal{S}}^2,
\\
\mathrm{Var}\left(\|\boldsymbol{V}_{\mathcal{S}}\boldsymbol{\phi}_{2i} - \boldsymbol{V}_{\mathcal{S}}\boldsymbol{\phi}_{2i-1}\|_2^2 \right)
\leq &~ \frac{12\sigma^4_{\mathrm{max}}\left(\mathcal{S}\right)}{K^3} \leq ~\frac{12d_{\mathcal{S}}^4}{K^3}
\label{Bounded difference property}
\end{align}
From the above inequalities we can see that $\|\boldsymbol{V}_{\mathcal{S}}\boldsymbol{\phi}_{2i} - \boldsymbol{V}_{\mathcal{S}}\boldsymbol{\phi}_{2i-1}\|_2^2 $ satisfies the Bernstein's condition with $b = 2d^2_{\mathcal{S}}$ (see eq. $2.15$ in \citep{wainwright2019high}). Then (from proposition $2.10$ in  \citep{wainwright2019high}) the following inequality holds with probability at least $1-\delta/6$, for any $\delta\in\left(0,1\right)$:
\begin{equation}
f_{\boldsymbol{V}_{\mathcal{S}}}\left(\boldsymbol{\phi}_1,\cdots, \boldsymbol{\phi}_{2m}\right) \geq \mathbb{E}\left[f_{\boldsymbol{V}_{\mathcal{S}}}\left(\boldsymbol{\phi}_1,\cdots, \boldsymbol{\phi}_{2m}\right)\right] - 2\sqrt{ \max\left\{\frac{4\log{\frac{6}{\delta}}}{m}, \frac{3}{K^3}\right\}\cdot \frac{d_{\mathcal{S}}^4}{m}\log{\frac{6}{\delta}}},
\label{Macdiarmid}
\end{equation}
where for the expected value of the function $f$ we have:
\begin{align}
\mathbb{E}\left[ f_{\boldsymbol{V}_{\mathcal{S}}}\left(\boldsymbol{\phi}_1,\cdots, \boldsymbol{\phi}_{2m}\right)\right] ~
= & ~ \frac{1}{2m}\sum_{i =1}^{m}{ \mathbb{E}\left[\|\boldsymbol{V}_{\mathcal{S}}\boldsymbol{\phi}_{2i} - \boldsymbol{V}_{\mathcal{S}}\boldsymbol{\phi}_{2i-1}\|_2^2\right]}
\nonumber \\ 
= & ~ \frac{1}{2} \mathbb{E}\left[\|\boldsymbol{V}_{\mathcal{S}}\boldsymbol{\phi} - \boldsymbol{V}_{\mathcal{S}}\boldsymbol{\phi}^\prime\|_2^2\right]
\nonumber \\  
= & ~ \frac{1}{2}\mathbb{E}\left[\left(\boldsymbol{V}_{\mathcal{S}}\boldsymbol{\phi} - \boldsymbol{V}_{\mathcal{S}}\boldsymbol{\phi}^\prime\right) ^T \left(\boldsymbol{V}_{\mathcal{S}}\boldsymbol{\phi} - \boldsymbol{V}_{\mathcal{S}}\boldsymbol{\phi}^\prime\right)\right]
\nonumber \\
= & ~ \frac{1}{2}\mathbb{E}\left[\left(\boldsymbol{\phi} -\boldsymbol{\phi}^\prime\right) ^T\boldsymbol{V}_{\mathcal{S}}^T\boldsymbol{V}_{\mathcal{S}} \left(\boldsymbol{\phi} - \boldsymbol{\phi}^\prime\right)\right]
\nonumber \\
= & ~ \frac{1}{2}\mathbb{E}\left[\Tr{\left(\boldsymbol{V}_{\mathcal{S}}^T\boldsymbol{V}_{\mathcal{S}} \left(\boldsymbol{\phi} - \boldsymbol{\phi}^\prime\right)\left(\boldsymbol{\phi} -\boldsymbol{\phi}^\prime\right) ^T\right)}\right]
\nonumber \\
= & ~ \frac{1}{2}\Tr{\left(\boldsymbol{V}_{\mathcal{S}}^T\boldsymbol{V}_{\mathcal{S}} \mathbb{E}\left[\left(\boldsymbol{\phi} - \boldsymbol{\phi}^\prime\right)\left(\boldsymbol{\phi} -\boldsymbol{\phi}^\prime\right)^T\right]\right)}
\nonumber \\
= & ~ \frac{1}{2}\Tr{\left(\boldsymbol{V}_{\mathcal{S}}^T\boldsymbol{V}_{\mathcal{S}}\frac{1}{(K+1)^2(K+2)}
  \begin{bmatrix}
    K & -1  & \dots  & -1 \\
    -1 & K  & \dots  & -1 \\
    \vdots  & \vdots  & \ddots & \vdots \\
    -1 & -1 & \dots  & K
\end{bmatrix}
\right)}
\nonumber \\
= & ~ \frac{1}{2(K+1)^2(K+2)}\Tr{\left(\boldsymbol{V}_{\mathcal{S}}^T\boldsymbol{V}_{\mathcal{S}}\left((K+1)\mathrm{\boldsymbol{I}} - 
  \boldsymbol{1}  \boldsymbol{1}^T\right)\right)}
\nonumber \\
= & ~ \frac{1}{2(K+1)^2(K+2)}\left((K+1)\Tr{\left(\boldsymbol{V}_{\mathcal{S}}^T\boldsymbol{V}_{\mathcal{S}}\right)} - \Tr{\left(
  \boldsymbol{1}^T\boldsymbol{V}_{\mathcal{S}}^T \boldsymbol{V}_{\mathcal{S}} \boldsymbol{1}  \right)}\right)
\nonumber \\
= & ~ \frac{1}{2(K+2)}\left(\sum_{i=1}^{K}{\frac{1}{K+1}\sum_{j=1}^{K+1}{\boldsymbol{V}_{\mathcal{S}}^{2}(i,j)}}- \sum_{i=1}^{K}{\left(\frac{1}{K+1}\sum_{j=1}^{K+1}{\boldsymbol{V}_{\mathcal{S}}(i,j)}\right)^2}\right)
\nonumber \\
= & ~ \frac{1}{2(K+2)}\sum_{i=1}^{K}{\frac{1}{K+1}\sum_{j=1}^{K+1}{\left(\boldsymbol{V}_{\mathcal{S}}(i,j) - \frac{1}{K+1}\sum_{j=1}^{K+1}{\boldsymbol{V}_{\mathcal{S}}(i,j)}\right)^2}}
\nonumber \\
= & ~ \frac{1}{2(K+2)}\sum_{i=1}^{K}{\frac{1}{K+1}\sum_{j=1}^{K+1}{\left(\widehat{\boldsymbol{V}}_{\mathcal{S}}(i,j) - \frac{1}{K+1}\sum_{j=1}^{K+1}{\widehat{\boldsymbol{V}}_{\mathcal{S}}(i,j)}\right)^2}}
\nonumber \\ 
= & ~ \frac{1}{2(K+2)}\sum_{i=1}^{K}{\frac{1}{K+1}\sum_{j=1}^{K+1}{\widetilde{\boldsymbol{V}}_{\mathcal{S}}(i,j)^2}}
\nonumber \\ 
= & ~ \frac{1}{2(K+2)(K+1)}\sum_{j=1}^{K+1}{\|\widetilde{\boldsymbol{V}}_{\mathcal{S}}^j\|_2^2}
\label{lowerboundForExpectation}
\end{align}
where $\widetilde{{\boldsymbol{V}}}_{\mathcal{S}}$ and $\widehat{{\boldsymbol{V}}}_{\mathcal{S}}$ are defined as
\begin{align}
\boldsymbol{V}_{\mathcal{S}} = &\left[\boldsymbol{v}_1,\boldsymbol{v}_2,\cdots,\boldsymbol{v}_{K+1}\right]
\\
\widehat{\boldsymbol{V}}_{\mathcal{S}} = &\left[\boldsymbol{0}, \boldsymbol{v}_2 - \boldsymbol{v}_1 ,\boldsymbol{v}_3 - \boldsymbol{v}_1 ,\cdots,\boldsymbol{v}_{K+1} - \boldsymbol{v}_1\right]
\nonumber
\\
= & \left[\hat{\boldsymbol{v}}_1,\hat{\boldsymbol{v}_2},\cdots,\hat{\boldsymbol{v}}_{K+1}\right]
\\
\widetilde{\boldsymbol{V}}_{\mathcal{S}} = &\left[\hat{\boldsymbol{v}}_1 - \frac{\sum_1^{K}{\hat{\boldsymbol{v}}_i}}{K+1} , \hat{\boldsymbol{v}}_2 - \frac{\sum_1^{K}{\hat{\boldsymbol{v}}_i}}{K+1} ,\cdots,\hat{\boldsymbol{v}}_K - \frac{\sum_1^{K}{\hat{\boldsymbol{v}}_i}}{K+1} \right]
\nonumber
\\
= & \left[\tilde{\boldsymbol{v}}_1,\tilde{\boldsymbol{v}_2},\cdots,\tilde{\boldsymbol{v}}_K,\tilde{\boldsymbol{v}}_{K+1}\right],
\end{align}
And $\widetilde{\boldsymbol{V}}_{\mathcal{S}}^j$ is the $j$th column of $\widetilde{\boldsymbol{V}}_{\mathcal{S}}$.
We know that the maximum distance between any two points inside a simplex is equal to the length of the biggest edge of the simplex. Now without loss of generality suppose that $\boldsymbol{v}_2$ and $\boldsymbol{v}_3$ are the vertices related to this edge (if more than one edge have the maximum length, suppose that, $\boldsymbol{v}_2$ and $\boldsymbol{v}_3$ are belong to one of them). Therefore the length of the vector $\boldsymbol{v}_3 - \boldsymbol{v}_2$ associated with the maximum distance satisfies the following
\begin{align}
    d_{\mathcal{S}}^2 ~= & \|\boldsymbol{v}_3 - \boldsymbol{v}_2\|_2^2
    \nonumber \\
    = & \|\tilde{\boldsymbol{v}_3} - \tilde{\boldsymbol{v}_2}\|_2^2
    \nonumber \\
    \leq & \|\tilde{\boldsymbol{v}_3}\|_2^2 + \|\tilde{\boldsymbol{v}_2}\|_2^2 + 2\|\tilde{\boldsymbol{v}_3}\|_2\|\tilde{\boldsymbol{v}_2}\|_2
    \nonumber \\
    \leq & 2\left(\|\tilde{\boldsymbol{v}}_3\|_2^2 + \|\tilde{\boldsymbol{v}}_2\|_2^2\right).
    \label{maximumLength}
\end{align}
Now the following inequality can be deduced from \ref{lowerboundForExpectation} and \ref{maximumLength}
\begin{equation}
    \mathbb{E}\left[ f_{\boldsymbol{V}_{\mathcal{S}}}\left(\boldsymbol{\phi}_1,\cdots, \boldsymbol{\phi}_{2m}\right)\right] \geq \frac{1}{4(K+1)(K+2)}d_{\mathcal{S}}^2.
    \label{lowerboundForExpectation2}
\end{equation}
From \ref{Macdiarmid} and \ref{lowerboundForExpectation2}, it is easy to show that if we have $2m \geq 2000(K+1)(K+2)\log{\frac{6}{\delta}}$ i.i.d. samples from $\mathbb{P}_{\mathcal{S}}$, then with probability at least $1-\frac{\delta}{6}$ (for any $0<\delta\leq 1$), we have
\begin{equation}
f\left(\boldsymbol{\phi}_1,\boldsymbol{\phi}_2,\cdots, \boldsymbol{\phi}_{2m}\right) \geq \frac{d_{\mathcal{S}}^2}{8(K+1)(K+2)}.
\label{Lowebound on emperical diameter}
\end{equation}
Next, we should find a lower-bound for the second term in the r.h.s. of \ref{Lower bound for statistic}, i.e., $\mathrm{II}$.

In \ref{Lower bound for statistic}, $\boldsymbol{z}_i$s are drawn from a multivariate Gaussian distribution with mean vector $\boldsymbol{0}$ and covariance matrix $\sigma^2 \boldsymbol{\mathrm{I}}$. Then, $\boldsymbol{z}^\prime_i = \tilde{\boldsymbol{z}}_{2i} - \tilde{\boldsymbol{z}}_{2i-1}$ is also a zero-mean Gaussian random vector with covariance matrix $\boldsymbol{\mathrm{I}}$, and as a result $\|\boldsymbol{z}^\prime_i\|_2^2$ is a Chi-squared random variable with $K$ degrees of freedom.  We know that a Chi-squared random variable with $K$ degrees of freedom is sub-exponential with parameters $\left(2\sqrt{K}, 4\right)$ and therefore $\frac{1}{2m}\sum_{i=1}^{m}{\|\boldsymbol{z}^\prime_i\|_2^2}$ is  a sub-exponential random variable with parameters $\left(\sqrt{{K}/{m}}, {2}/{m}\right)$. Then the following holds with probability at least $1-\delta/6$:
\begin{align}
\frac{1}{2m}\sum_{i=1}^{m}{\|\tilde{\boldsymbol{z}}_{2i}-\tilde{\boldsymbol{z}}_{2i-1}\|_2^2}~
\geq&~ \frac{1}{2m}\sum_{i=1}^{m}{\mathbb{E}\left[\|\tilde{\boldsymbol{z}}_{2i}-\tilde{\boldsymbol{z}}_{2i-1}\|_2^2\right]} - \sqrt{\frac{2K}{m}\log{\frac{6}{\delta}}}
\nonumber \\
\geq & ~ \frac{K}{2}-\sqrt{\frac{2K}{m}\log{\frac{6}{\delta}}}
\nonumber \\
\geq &~ \frac{K}{2}-\sqrt{\frac{K}{1000(K+1)(K+2)}}
\nonumber \\
\geq &~ \frac{K}{2} - 1 = \frac{K-2}{2}.
\end{align}
Based on the above inequalities, we can give a lower-bound for the second term in the r.h.s. of \ref{Lower bound for statistic}:
\begin{equation}
\mathrm{II} = \frac{2\sigma^2}{2m}\sum_{i=1}^{m}{\|\tilde{\boldsymbol{z}}_{2i} - \tilde{\boldsymbol{z}}_{2i-1} \|_2^2} \geq (K-2)\sigma^2.
\label{upper bound subexponential}
\end{equation}
Now to find a lower bound for $\mathrm{D}\left(\mathrm{S}\right)$, we only need to give a lower bound for $\mathrm{III}$, in the last inequality of \ref{Lower bound for statistic}. For this part we have:
\begin{align}
\mathrm{III} 
~= & ~\frac{\sigma\sqrt{2}}{m}\sum_{i=1}^{m}{\left(\boldsymbol{V}_{\mathcal{S}}\boldsymbol{\phi}_{2i} - \boldsymbol{V}_{\mathcal{S}}\boldsymbol{\phi}_{2i-1}\right)^T\left(\tilde{\boldsymbol{z}}_{2i} - \tilde{\boldsymbol{z}}_{2i-1}\right)}
\nonumber \\
= & ~ \frac{\sigma\sqrt{2}}{m}\sum_{i=1}^{m}{\left(\boldsymbol{V}_{\mathcal{S}}\boldsymbol{\phi}_{2i} - \boldsymbol{V}_{\mathcal{S}}\boldsymbol{\phi}_{2i-1}\right)^T\boldsymbol{z}_i^\prime}
\nonumber \\
= & ~ \frac{\sigma\sqrt{2}}{m}\sum_{i=1}^{m}{g_{\boldsymbol{\phi_{2i}}, \boldsymbol{\phi_{2i-1}}}\left(z_i^{\prime1}, z_i^{\prime2},\cdots, z_i^{\prime m}\right)},
\label{lower bound for 3rd part}
\end{align}
where in the above equations we have:
 \begin{equation}
  \mathbb{E}\left(g_{\boldsymbol{\phi_{2i}}, \boldsymbol{\phi_{2i-1}}}\left(z_i^{\prime1}, z_i^{\prime2},\cdots, z_i^{\prime m}\right)\right) 
   = \mathbb{E}\left(\boldsymbol{V}_{\mathcal{S}}\boldsymbol{\phi}_{2i} - \boldsymbol{V}_{\mathcal{S}}\boldsymbol{\phi}_{2i-1}\right)^T\mathbb{E}\left(\boldsymbol{z}_i^\prime\right) = 0.
 \end{equation}
To find a lower bound for $\mathrm{III}$ we try to bound $g_{\boldsymbol{\phi_{2i}}, \boldsymbol{\phi_{2i-1}}}\left(z_i^{\prime1}, z_i^{\prime2},\cdots, z_i^{\prime m}\right)$ from below. To do so, we write the concentration inequality for this function as follow:
\begin{align}
\mathbb{P}\left(g_{\boldsymbol{\phi_{2i}}, \boldsymbol{\phi_{2i-1}}}\left(z_i^{\prime1}, z_i^{\prime2},\cdots, z_i^{\prime m}\right) \geq \epsilon\right)
\leq & ~ \frac{\mathbb{E}_{\boldsymbol{\phi_{2i}}, \boldsymbol{\phi_{2i-1}}, \boldsymbol{z}_i^\prime}\left[e^{\lambda g_{\boldsymbol{\phi_{2i}}, \boldsymbol{\phi_{2i-1}}}\left(z_i^{\prime1}, z_i^{\prime2},\cdots, z_i^{\prime m}\right) }\right]}{e^{\lambda\epsilon}}
\nonumber \\
\leq & ~ \frac{\mathbb{E}_{\boldsymbol{\phi_{2i}}, \boldsymbol{\phi_{2i-1}}, \boldsymbol{z}_i^\prime}\left[e^{\lambda g_{\boldsymbol{\phi_{2i}}, \boldsymbol{\phi_{2i-1}}}\left(z_i^{\prime1}, z_i^{\prime2},\cdots, z_i^{\prime m}\right) }\right]}{e^{\lambda\epsilon}}
\nonumber \\
= & ~ \frac{\mathbb{E}_{\boldsymbol{\phi_{2i}}, \boldsymbol{\phi_{2i-1}}}\left[\mathbb{E}_{\boldsymbol{z}_i^\prime}\left[e^{\lambda g_{\boldsymbol{\phi_{2i}}, \boldsymbol{\phi_{2i-1}}}\left(z_i^{\prime1}, z_i^{\prime2},\cdots, z_i^{\prime m}\right) }\right]\right]}{e^{\lambda\epsilon}},
\label{concentration for g}
\end{align}
where $g_{\boldsymbol{\phi_{2i}}, \boldsymbol{\phi_{2i-1}}}\left(\boldsymbol{z}\right)$ is a lipschitz function with respect to $\boldsymbol{z}^\prime$:
\begin{align}
\vert g_{\boldsymbol{\phi_{2i}}, \boldsymbol{\phi_{2i-1}}}\left(\boldsymbol{z}_1\right) - g_{\boldsymbol{\phi_{2i}}, \boldsymbol{\phi_{2i-1}}}\left(\boldsymbol{z}_2\right)\vert
\leq& \quad \|\boldsymbol{V}_{\mathcal{S}}\boldsymbol{\phi}_{2i} - \boldsymbol{V}_{\mathcal{S}}\boldsymbol{\phi}_{2i-1}\|_2\|\boldsymbol{z}_1 - \boldsymbol{z}_2\|_2
\nonumber \\
\leq& \quad 2d_\mathcal{S}\|\boldsymbol{z}_1 - \boldsymbol{z}_2\|_2.
\end{align}
From Lemma (2.27) in \citep{wainwright2019high} we know that any $L$-lipschitz function of a Gaussian R.V. is sub-Gaussian. As a result, we have
\begin{align}
\mathbb{E}_{\boldsymbol{z}_i^\prime}\left[e^{\lambda g_{\boldsymbol{\phi_{2i}}, \boldsymbol{\phi_{2i-1}}}\left(z_i^{\prime1}, z_i^{\prime2},\cdots, z_i^{\prime m}\right) }\right]
\leq & \quad e^{\frac{4\pi^2d_{\mathcal{S}}^2\lambda^2}{8}}.
\label{upperbound for mgf of L-lipschitz}
\end{align}
From inequalities \ref{upperbound for mgf of L-lipschitz}, \ref{concentration for g} and equation \ref{lower bound for 3rd part} we have:
\begin{equation}
\mathbb{P}\left(\frac{\sigma\sqrt{2}}{2m}\sum_{i=1}^{m}{g_{\boldsymbol{\phi_{2i}}, \boldsymbol{\phi_{2i-1}}}\left(\boldsymbol{z}_i\right)} \leq -\epsilon\right)
\leq \min_{\lambda \geq 0} \frac{\mathbb{E}_{\boldsymbol{\phi_{1}},\cdots \boldsymbol{\phi_{2m}}}\left[e^{\frac{\lambda^2 \pi^2 d_{\mathcal{S}}^2\sigma^2}{2m}\cdot}\right]}{e^{\lambda\epsilon}}
\leq e^{-\frac{m\epsilon^2}{2\pi^2d_{\mathcal{S}}^2\sigma^2}}.
\end{equation}
Then if we have $2m \geq 2000(K+1)(K+2)\log{\frac{6}{\delta}}$, i.i.d samples, then with probability at least $1-\frac{\delta}{6}$ we have:
\begin{equation}
\frac{1}{2m}\sum_{i=1}^{m}{\left(\boldsymbol{V}_{\mathcal{S}}\boldsymbol{\phi}_{2i} - \boldsymbol{V}_{\mathcal{S}}\boldsymbol{\phi}_{2i-1}\right)^T\left(\boldsymbol{z}_{2i} - \boldsymbol{z}_{2i-1}\right)} \geq -\frac{\pi\sigma d_{\mathcal{S}}}{20\sqrt{(K+2)(K+1)}} .
\label{lower bound for inner product}
\end{equation}
Now from inequalities  \ref{Lowebound on emperical diameter},\ref{upper bound subexponential}, \ref{lower bound for inner product} and equation \ref{Lower bound for statistic} we have: 
\begin{align}
\mathrm{D} 
\geq &  \frac{d_{\mathcal{S}}^2}{8(K+1)(K+2)} + (K-2) \sigma^2 - \frac{\pi\sigma d_{\mathcal{S}}}{20\sqrt{(K+2)(K+1)}}
\nonumber \\
= &  \frac{d_{\mathcal{S}}^2}{8(K+1)(K+2)} + \left(\sigma\sqrt{K-2} - \frac{d_{\mathcal{S}}\sqrt{\pi}}{40\sqrt{\left(K+1\right)\left(K+2\right)\left(K-2\right)}}\right)^2 -
\nonumber \\ &
\frac{d_{\mathcal{S}}^2\pi}{1600\left(K+1\right)\left(K+2\right)\left(K-2\right)} 
\label{upper bound for noise varaince}
\\ 
=& (K-2) \sigma^2 + \left(\frac{\pi\sigma}{10\sqrt{2}} - \frac{d_{\mathcal{S}}}{2\sqrt{2}\sqrt{(K+1)(K+2)}}\right)^2 - \frac{\pi^2\sigma^2}{200} .
\label{upper bound for maximum singular value}
\end{align}
From the inequalities \ref{upper bound for maximum singular value} and \ref{upper bound for noise varaince} it can be seen that if we have $2m \geq 2000(K+1)(K+2)\log{\frac{6}{\delta}}$ then the following inequalities with probability at least $1-\delta/2$ give upper-bounds for the maximum distance between two points in the simplex $\mathcal{S}$, and the noise variance $\sigma$:
\begin{align}
d_{\mathcal{S}} \leq & 4\sqrt{(K+1)(K+2)D} = R.
\label{Radius}
\\
\sigma^2 \leq & \frac{D}{K-3} = R_n
\label{noise radius}
\end{align}

In the same way we could also give an upper bound for $\mathrm{D}$. The following inequalities hold with probability more than $1- \delta/2$:
\begin{align}
\mathrm{D}
\leq &  \frac{d_{\mathrm{S}}^2}{K+2} + \sigma^2(K+2) +\frac{\pi\sigma d_{\mathcal{S}}}{20\sqrt{(K+2)(K+1)}} 
\nonumber \\
\leq & \left(\frac{d_{\mathrm{S}}}{\sqrt{K+2}} + \sigma\sqrt{K+2}\right)^2
\nonumber \\
 = & \frac{d_{\mathrm{S}}^2}{K+2}\left(1+\frac{K+2}{d_{\mathrm{S}}/\sigma}\right)^2
\label{upperbound for empirical diameter}
\\
= & \sigma^2\left(K+2\right)\left(1+\frac{d_{\mathrm{S}}/\sigma}{K+2}\right)^2.
\label{upperbound for empirical noise variance}
\end{align}
Based on equation \ref{Radius} and inequations in \ref{upperbound for empirical diameter} and \ref{upperbound for empirical noise variance}, we can give an upper bound for the radius $R$ in \ref{Radius} and noise radius $R_n$. The following inequality holds with probability at least $1-\delta$:
\begin{align}
R \leq & 4\sqrt{K+1}\left(1+\frac{K+2}{d_{\mathrm{S}}/\sigma}\right)d_{\mathrm{S}}
\label{final radius of containing sphere}
\\
R_n \leq & \frac{K+2}{K-3}\left(1+\frac{d_{\mathrm{S}}/\sigma}{K+2}\right)\sigma.
\label{final upperbound of noise variance}
\end{align}

Now the only thing we should do is to find a point inside of the simplex. To do this we define a new statistic $\mathrm{\boldsymbol{p}}$ as follows:
\begin{align}
\mathrm{\boldsymbol{p}} 
&= \frac{1}{2m}\sum_{i=1}^{2m}{\boldsymbol{x}_i} 
= \frac{1}{2m}\sum_{i=1}^{2m}{\boldsymbol{V}_{\mathcal{S}}\boldsymbol{\phi}_{i}+\boldsymbol{z}_i} 
= \boldsymbol{V}_{\mathcal{S}}\left( \frac{1}{2m}\sum_{i=1}^{2m}{\boldsymbol{\phi}_{i}}\right)+\frac{1}{2m}\sum_{i=1}^{2m}{\boldsymbol{z}_i} 
 = \mathrm{I} + \mathrm{II}
\label{center point statistic}
\end{align}
It is clear that `'I'' is placed inside of the true simplex. So the distance between $\mathrm{\boldsymbol{p}} $ and a point inside the main simplex can be calculated as follows:
\begin{align}
\min_{\boldsymbol{x} \in \mathcal{S}}{\left\Vert\mathrm{\boldsymbol{p}} -\boldsymbol{x}\right\Vert_2}
& \leq \left\Vert \frac{1}{2m}\sum_{i=1}^{2m}{\boldsymbol{z}_i} \right\Vert_2.
\end{align}

In the above equation $\frac{1}{2m}\sum_{i=1}^{2m}{\boldsymbol{z}_i}$, is a zero mean Gaussian random vector with covariance matrix $\frac{\sigma^2}{2m}\boldsymbol{I}$, therefore $ \| \frac{1}{2m}\sum_{i=1}^{2m}{\boldsymbol{z}_i} \|_2^2$ is a chi-squared random variable with $K$ degrees of freedom, and so we can write concentration inequality for it:
\begin{align}
\mathrm{P}\left[\left\vert \left\Vert \frac{1}{2m}\sum_{i=1}^{2m}{\boldsymbol{z}_i} \right\Vert_2^2 - \frac{K\sigma^2}{2m}\right\vert \geq \epsilon \right] \leq 2e^{-\frac{m^2\epsilon^2}{2K\sigma^4}}.
\end{align}
 Then with probability at least $1-\delta$ we have the following inequality:
 \begin{align}
\min_{\boldsymbol{x} \in \mathcal{S}}{\|\mathrm{\boldsymbol{p}} -\boldsymbol{x}\|_2^2}
\leq &\quad \frac{K\sigma^2}{2m} + \frac{\sigma^2}{m}\sqrt{2K\log{\frac{2}{\delta}}}
\nonumber \\
\leq &\quad \frac{K\sigma^2}{m}\log{\frac{2}{\delta}} 
\end{align}

Now if we have $2m \geq 2000(K+1)(K+2)\log{\frac{6}{\delta}}$, then we can rewrite the above inequality as follows:
\begin{align}
\min_{\boldsymbol{x} \in \mathcal{S}}{\|\boldsymbol{p} -\boldsymbol{x}\|_2}
\leq \quad \frac{\sigma}{10\sqrt{10}\sqrt{K+2}}.
\end{align}

As we mentioned earlier, the first part in the right hand side of the inequality \ref{center point statistic}  , $\frac{1}{2m}\sum_{i=1}^{2m}{\boldsymbol{V}_{\mathcal{S}}\boldsymbol{\phi}_{i}}$, belongs to the interior of the simplex. Then a $K$-dimensional sphere with radius $R$ and centered at this point, with probability at least $1 - \delta$ contains the simplex $\mathcal{S}$. However, in the process of learning we do not have access to the noiseless data and thus cannot have such a point as the center of the sphere. Let us show the distance between $\boldsymbol{p}$ and $\frac{1}{2m}\sum_{i=1}^{2m}{\boldsymbol{V}_{\mathcal{S}}}$ with $d$. Then, it is clear to see that the $K$-dimensional sphere with radius $R+d$ and center point $\boldsymbol{p}$, contains the sphere with center point at $\frac{1}{2m}\sum_{i=1}^{2m}{\boldsymbol{V}_{\mathcal{S}}\boldsymbol{\phi}_{i}}$ and radius $R$. So any simplex which is placed in sphere $\mathrm{C}^{K}(\frac{1}{2m}\sum_{i=1}^{2m}{\boldsymbol{V}_{\mathcal{S}}\boldsymbol{\phi}_{i}}, R)$ is also placed in sphere $\mathrm{C}^{K}(\boldsymbol{p}, R+d)$.

So we conclude that if we have $2m \geq 2000(K+1)(K+2)\log{\frac{6}{\delta}}$ i.i.d. samples from $\mathbb{G}_{\mathcal{S}}$ then the main simplex $\mathcal{S}$ with probability more than or equal to $1- \delta$ will be confined in a $K$-dimensional sphere with center point at $\boldsymbol{p}$ and with radius $R$:
\begin{align}
R = ~& 4\sqrt{(K+1)(K+2)D} + \frac{\sqrt{D}}{40\sqrt{K+2}} 
\nonumber\\
\leq &~ 4\sqrt{(K+1)(K+2)D}\left(1+\frac{1}{160(K+2)\sqrt{K+1}}\right)
\nonumber \\
\leq &~ 8\sqrt{(K+1)(K+2)D}.
\label{final upperbound}
\end{align}
This completes the proof.
\end{proof}

\vspace*{3mm}
\begin{proof}[proof of Lemma \ref{quantization lemma}]
Consider a $\left(\underline{\theta},\bar{\theta}\right)$-isoperimetric $K$-simplex $\mathcal{S}$, which is bounded in a sphere $\mathrm{C}^{K}(\boldsymbol{p}, R)$. It is clear that all vertices of this simplex $\left\{\boldsymbol{v}_1,\ldots,\boldsymbol{v}_{K+1}\right\}$ are placed in $\mathrm{C}^{K}(\boldsymbol{p}, R)$. From the definition of $\mathrm{T}_{\frac{\alpha\epsilon}{K+1}}(\mathrm{C}^{K}(\boldsymbol{p}, R))$, we know that for each vertex of the simplex $\boldsymbol{v}_i$, there exists some $\boldsymbol{v}_i^{\prime} \in \mathrm{T}_{\frac{\alpha\epsilon}{K+1}}(\mathrm{C}^{K}(\boldsymbol{p}, R))$ such that $\|\boldsymbol{v}_i - \boldsymbol{v}_i^{\prime}\|_2 \leq \frac{\alpha\epsilon}{K+1}$. Assume for each vertex $\boldsymbol{v}_i$ of $\mathcal{S}$, we denote its closest point in $ \mathrm{T}_{\frac{\alpha\epsilon}{K+1}}(\mathrm{C}^{K}(\boldsymbol{p}, R))$ as
\begin{equation*}
\hat{\boldsymbol{v}}_i = \argmin \left\{\|\boldsymbol{v}_i - \hat{\boldsymbol{v}}\|_2  \bigg\vert ~\hat{\boldsymbol{v}} \in  \mathrm{T}_{\frac{\alpha\epsilon}{K+1}}(\mathrm{C}^{K}(\boldsymbol{p}, R))\right\}.
\end{equation*}
Using these points we make a new simplex $\widehat{\mathcal{S}}$. It is clear that $\widehat{\mathcal{S}}$ belongs to $\widehat{\mathbb{S}}\left(\mathrm{C}^{K}(\boldsymbol{p}, R)\right)$.
Assume $f_{\mathcal{S}}$ and $f_{\hat{\mathcal{S}}}$ denote the probability density functions that correspond to $\mathcal{S}$ and $\widehat{\mathcal{S}}$, respectively. Then, the TV-distance between $f_{\mathcal{S}}$ and $f_{\widehat{\mathcal{S}}}$ can be written and bounded as
\begin{align}
\mathrm{TV}\left(\mathbb{P}_{\mathcal{S}}, \mathbb{P}_{\widehat{\mathcal{S}}}\right) 
=&~ \sup_{A}~{\mathbb{P}_{\mathcal{S}}\left(A\right)- \mathbb{P}_{\widehat{\mathcal{S}}}\left(A\right)}
\nonumber \\
=& \int_{\boldsymbol{x} \in \{x^\prime:f_{\mathcal{S}}\left(\boldsymbol{x}^\prime\right) \ge f_{\widehat{\mathcal{S}}}\left(\boldsymbol{x}^\prime\right)\}}{f_{\mathcal{S}}\left(\boldsymbol{x}\right) - f_{\widehat{\mathcal{S}}}\left(\boldsymbol{x}\right)}
    \nonumber \\ 
    =& \int_{\boldsymbol{x} \in \mathcal{S}-\widehat{\mathcal{S}}}{f_{\mathcal{S}}\left(\boldsymbol{x}\right) - f_{\widehat{\mathcal{S}}}\left(\boldsymbol{x}\right)} + \left(\int_{\boldsymbol{x} \in \mathcal{S}\cap\widehat{\mathcal{S}}}{f_{\mathcal{S}}\left(\boldsymbol{x}\right) - f_{\widehat{\mathcal{S}}}\left(\boldsymbol{x}\right)}\right)\boldsymbol{\mathrm{1}}\left(\mathrm{Vol}\left(\mathcal{S}\right) \leq \mathrm{Vol}\left(\widehat{\mathcal{S}}\right)\right)
    \nonumber \\
    \leq& \int_{\boldsymbol{x} \in \mathcal{S}-\widehat{\mathcal{S}}}{f_{\mathcal{S}}\left(\boldsymbol{x}\right) - f_{\widehat{\mathcal{S}}}\left(\boldsymbol{x}\right)} + \int_{\boldsymbol{x} \in \mathcal{S}\cap\widehat{\mathcal{S}}}{\bigg\vert f_{\mathcal{S}}\left(\boldsymbol{x}\right) - f_{\widehat{\mathcal{S}}}\left(\boldsymbol{x}\right)\bigg\vert}
    \nonumber \\
    =& \frac{\mathrm{Vol}\left(\mathcal{S-\widehat{S}}\right)}{\mathrm{Vol}\left(S\right)} + \mathrm{Vol}\left( \mathcal{S} \cap \widehat{\mathcal{S}} \right)\bigg\vert\frac{1}{\mathrm{Vol}(\mathcal{S})} - \frac{1}{\mathrm{Vol}(\widehat{\mathcal{S}})}\bigg\vert,
    \label{upper bound for tv after quantization}
\end{align}
where $\mathcal{S} - \widehat{\mathcal{S}}$ shows the set difference between $\mathcal{S}$ and $\widehat{\mathcal{S}}$. Let us denote the two terms in the r.h.s. of \ref{upper bound for tv after quantization} as $\mathrm{I}$ and $\mathrm{II}$, respectively. To find an upper-bound for the $\mathrm{TV}$-distance in \ref{upper bound for tv after quantization} we  find respective upper-bounds for $\mathrm{I}$ and $\mathrm{II}$. In order to do so, first let us discuss about $\mathrm{I}$:
\begin{align}
\mathrm{I} = \frac{\mathrm{Vol}\left(\mathcal{S-\widehat{S}}\right)}{\mathrm{Vol}\left(S\right)}. 
\end{align}
The distance between vertices of $\mathcal{S}$ and $\widehat{\mathcal{S}}$ are less than $\frac{\alpha\epsilon}{K+1}$, then the maximal difference set (in terms of volume) between $\mathcal{S}$ and $\widehat{\mathcal{S}}$ can be bounded as follows: simplex has $K+1$ facets, and the difference set that can occur from altering each of them is upper-bounded as $\leq \mathcal{A}_i\left(\mathcal{S}\right)\times \alpha\epsilon/(K+1)$, where $i=1,\ldots,K+1$ denotes the facet index. This way, for $\mathrm{I}$ we have:
\begin{align}
\mathrm{I} 
\leq&~\sum_{i=1}^{K+1}{\frac{\alpha\epsilon}{K+1}\frac{\mathcal{A}_i\left(\mathcal{S}\right)}{\mathrm{Vol}\left(\mathcal{S}\right)}}
\nonumber \\
\leq&~\alpha\epsilon\mathcal{A}_{max}\left(\mathcal{S}\right)
\nonumber \\ 
\leq&~\alpha\epsilon\bar{\theta}\mathrm{Vol}^{\frac{1}{K}},
\label{upper bound for tv after quantization part I}
\end{align}
where $\mathcal{A}_{max}\left(\mathcal{S}\right)$ is the volume of the largest facet of $\mathcal{S}$. In the final inequality of \ref{upper bound for tv after quantization part I}, we take advantage of the isoperimetricity property of $\mathcal{S}$. Next, we should find an upper-bound for the second term in the r.h.s. of \ref{upper bound for tv after quantization}, i.e., $\mathrm{II}$:
\begin{equation}
\mathrm{II} = \mathrm{Vol}\left( \mathcal{S} \cap \widehat{\mathcal{S}} \right)\bigg\vert\frac{1}{\mathrm{Vol}(\mathcal{S})} - \frac{1}{\mathrm{Vol}(\widehat{\mathcal{S}})}\bigg\vert.
\end{equation}
To find an upper-bound for $\mathrm{II}$ we should find an upper-bound for $\mathrm{Vol}\left(\widehat{\mathcal{S}}\right)$. To do so, we create a ``rounded" simplex $\mathcal{S}^r$ which forms by adding a $K$-dimensional sphere with radius $r = \frac{\alpha\epsilon}{K+1}$ to the original simplex $\mathcal{S}$. It can be shown that $\widehat{\mathcal{S}}$ is definitely placed inside $\mathcal{S}^r$ and therefore $\mathrm{Vol}(\hat{\mathcal{S}}) \leq \mathrm{Vol}(\mathcal{S}^r)$. In this regard, we have
\begin{align}
\mathrm{II} 
=&~ \mathrm{Vol}\left( \mathcal{S} \cap \widehat{\mathcal{S}} \right)\bigg\vert\frac{1}{\mathrm{Vol}(\mathcal{S})} - \frac{1}{\mathrm{Vol}(\hat{\mathcal{S}})}\bigg\vert
\nonumber \\
\leq&~ \bigg\vert\frac{\mathrm{Vol}(\hat{\mathcal{S}}) - \mathrm{Vol}(\mathcal{S})}{\mathrm{Vol}(\mathcal{S})}\bigg\vert
\nonumber \\
\leq&~ \bigg\vert\frac{\mathrm{Vol}(\mathcal{S}^r) - \mathrm{Vol}(\mathcal{S})}{\mathrm{Vol}(\mathcal{S})}\bigg\vert
\nonumber \\
\leq&~ \left(\sum_{i=1}^{K+1}{\frac{\alpha\epsilon}{K+1}\mathcal{A}_i\left(\mathcal{S}\right)} + (K+1)\mathcal{C}_{\alpha}\left(\frac{\alpha\epsilon}{K+1}\right)^K\right)\frac{1}{\mathrm{Vol}(\mathcal{S})}
\nonumber \\
\leq&~ \left((K+1)\frac{\alpha\epsilon}{K+1}\mathcal{A}_{\max}\left(\mathcal{S}\right) + (K+1)\mathcal{C}_{\alpha}\left(\frac{\alpha\epsilon}{K+1}\right)^K\right)\frac{1}{\mathrm{Vol}(\mathcal{S})}
\nonumber \\
\leq&~ 2\alpha\epsilon\bar{\theta}\mathrm{Vol}^{\frac{1}{K}},
\label{upper bound for tv after quantization part II}
\end{align}
where $\mathcal{C}_{\alpha}\left(\frac{\alpha\epsilon}{K+1}\right)^K$ is the volume of a $K$-dimensional sphere with radius $r = \frac{\alpha\epsilon}{K+1}$. Now, using \ref{upper bound for tv after quantization part I} and \ref{upper bound for tv after quantization part II} we have
\begin{align}
\mathrm{TV}\left(\mathbb{P}_{\mathcal{S}}, \mathbb{P}_{\widehat{\mathcal{S}}}\right)
\leq 3\alpha\epsilon\bar{\theta}\mathrm{Vol}^{\frac{1}{K}} \leq \frac{3}{5}\epsilon \leq \epsilon.
\label{upper bound for tv after quantization final}
\end{align}
From \ref{upper bound for tv after quantization final}, it can be seen that for any $\left(\underline{\theta},\bar{\theta}\right)$-isoperimetric simplex $\mathcal{S}$ which is bounded in a $K$-dimensional sphere $\mathrm{C}^{K}(\boldsymbol{p}, R)$, there exists at least one simplex in $\widehat{\mathbb{S}}(\mathrm{C}^{K}(\boldsymbol{p}, R))$ within a TV-distance of $\epsilon$ from $\mathcal{S}$. Thus the proof is complete.

In the end, it is worth mentioning one possible method to build an $\epsilon$-covering set for a $K$-dimensional sphere $\mathrm{C}^{K}(\boldsymbol{p}, R)$. Although there exists deterministic ways to build this set, an alternative approach is by uniformly sampling points from the sphere. Consider an $\epsilon/2$-packing set with size $L$ for the sphere. Based on the result of the ``coupon collector problem" we know that having $L\log(L)$ uniform samples from the sphere guarantees with high probability that there exists at least one sample within a distance of $\epsilon/2$ from each element of the packing set. Therefore, the resulting samples are an $\epsilon$-covering set for the sphere. Hence, the cardinality of an $\epsilon$-covering set built in this way is at most $\left(1+\frac{4R}{\epsilon}\right)^{2K}$.
\end{proof}


\section{Consistency Analysis}

Assume $\mathcal{S}_1,\mathcal{S}_2\in\mathbb{S}_K$ represent two arbitrary simplices in $\mathbb{R}^K$. In this regard, let $\mathbb{P}_{\mathcal{S}_1}$ and $\mathbb{P}_{\mathcal{S}_2}$ denote the probability measures, and $f_{\mathcal{S}_1}$ and $f_{\mathcal{S}_2}$ represent the probability density functions associated to $\mathcal{S}_1$ and $\mathcal{S}_2$, respectively. Let us assume that $\mathcal{S}_1,\mathcal{S}_2$ have a minimum degree of geometric regularity in the following sense.

\begin{definition}
For a simplex $\mathcal{S}\in\mathbb{S}_K$ with vertices $\boldsymbol{\theta}_0,\ldots,\boldsymbol{\theta}_{K}\in\mathbb{R}^K$, we say $\mathcal{S}$ is $\left(\bar{\lambda},\underline{\lambda}\right)$-regular if
$$
\underline{\lambda}
\leq
\lambda_{\min}\left(\boldsymbol{\Theta}\right)
\leq
\lambda_{\max}\left(\boldsymbol{\Theta}\right)
\leq
\bar{\lambda},
$$
where $\lambda_{\max}\left(\cdot\right)$ and $\lambda_{\min}\left(\cdot\right)$ denotes the largest and smallest eigenvalues of a matrix, respectively. Here, $\boldsymbol{\Theta}$ represents the zero-centerd vertex matrix of $\mathcal{S}$, i.e.,
$$
\boldsymbol{\Theta}\triangleq
\left[
\boldsymbol{\theta}_1-\boldsymbol{\theta}_0
\vert
\cdots
\vert
\boldsymbol{\theta}_K-\boldsymbol{\theta}_0
\right].
$$
\end{definition}
This definition has tight connections to the previously used notion of $\left(\bar{\theta},\underline{\theta}\right)$-isoperimetricity which is already used in the main body of the manuscript. In fact, it can be shown that $\mathcal{L}_{\max}=\mathcal{O}\left(\Bar{\lambda}\right)$ and vice versa.


Our aim is to show that if the noisy versions of $\mathcal{S}_1$ and $\mathcal{S}_2$, which we denote by $f_{\mathcal{S}_1}*G_{\sigma}$ and $f_{\mathcal{S}_2}*G_{\sigma}$, respectively, have a maximum total variation distance of at least $\epsilon>0$, then the TV distance between $\mathbb{P}_{\mathcal{S}_1}$ and $\mathbb{P}_{\mathcal{S}_2}$ is also bounded away from zero according to a function of $\epsilon,\sigma$ and the geometric reqularity of simplices $\mathcal{S}_1$ and $\mathcal{S}_2$. The theoretical core behind our method is stated in the following general theorem.

\begin{theorem}[Recovery of Low-Frequency Objects from Additive Noise]
\label{thm:generalResultTheorem-Appendix}
For $K\in\mathbb{N}$, consider a probability density function family $\mathscr{F}\subseteq\mathcal{M}\left(\mathbb{R}^K\right)$, i.e., a subset of distributions supported over $\mathbb{R}^K$. Assume for sufficiently large $\alpha>0$, the following bound holds for all $f,g\in\mathscr{F}$:
\begin{equation*}
\frac{1}{\left(2\pi\right)^K}
\int_{\left\Vert\boldsymbol{\omega}\right\Vert_{\infty}\ge\alpha}
\left\vert
\mathcal{F}\left\{f\right\}\left(\boldsymbol{\omega}\right)
-
\mathcal{F}\left\{g\right\}\left(\boldsymbol{\omega}\right)
\right\vert^2
\leq
\zeta\left(\alpha^{-1}\right)
\int_{\mathbb{R}^K}
\left\vert
f-g
\right\vert^2,
\end{equation*}
where $\mathcal{F}\left\{\cdot\right\}$ denotes the Fourier transform, and $\zeta$ is an increasing fuction with $\zeta\left(0\right)=0$ and continuity at $0$. Also, assume the probability density function $Q\in\mathcal{M}\left(\mathbb{R}^K\right)$ again for a sufficiently large $\alpha>0$ has the following property:
\begin{equation*}
\inf_{\left\Vert\boldsymbol{\omega}\right\Vert_{\infty}\leq\alpha}
\left\Vert
\mathcal{F}\left\{Q\right\}\left(\boldsymbol{\omega}\right)
\right\Vert
\ge \eta\left(\alpha\right),
\end{equation*}
where $\eta\left(\cdot\right)$ is a non-negative decreasing function. Then, there exists a non-negative constant $C$ where for any $\sigma,\varepsilon>0$ and $f,g\in\mathscr{F}$ with
$
\left\Vert
f-g
\right\Vert_2
\geq\varepsilon,
$
we have
$$
\left\Vert
\left(f-g\right)*Q
\right\Vert_2
\geq
\frac{\varepsilon}{\left(2\pi\right)^K}
\left(
\sup_{\alpha>C}~
\eta\left(\alpha\right)
\sqrt{1-\zeta\left(\alpha^{-1}\right)}
\right),
$$
with $*$ denoting the multi-dimensional convolution operator.
\end{theorem}
\begin{proof}
For the sake of simplicity in notations, let $\mathcal{F},\mathcal{G},\mathcal{Q}:\mathbb{R}^K\rightarrow\mathbb{C}$ denote the Fourier transforms of $f,g$ and $Q$, respectively. Due to Parseval's theorem, we have
$$
\left\Vert
f-g
\right\Vert^2_2
=
\frac{1}{\left(2\pi\right)^K}
\left\Vert
\mathcal{F}-\mathcal{G}
\right\Vert^2_2.
$$
Also, due to the properties of the Fourier transform, which is the transofrmation of convolution into direct multiplication, one can write
$$
\mathcal{F}\left\{
\left(f-g\right)*Q
\right\}
=
\mathcal{Q}\left(\mathcal{F}-\mathcal{G}\right).
$$
Thus, there exists universal constant $C>0$ such that for any $\alpha>C$:
\begin{align}
\left(2\pi\right)^K
\left\Vert
\left(f-g\right)*Q
\right\Vert^2_2
&=
\int_{\mathbb{R}^K}
\left\vert
\mathcal{Q}
\left(\boldsymbol{\omega}\right)
\left(
\mathcal{F}\left(\boldsymbol{\omega}\right)
-
\mathcal{G}\left(\boldsymbol{\omega}\right)
\right)
\right\vert^2
\\
&\ge
\int_{\left\Vert\boldsymbol{\omega}\right\Vert_{\infty}\leq\alpha}
\left\vert
\mathcal{Q}
\left(\boldsymbol{\omega}\right)
\right\vert^2
\left\vert
\mathcal{F}\left(\boldsymbol{\omega}\right)
-
\mathcal{G}\left(\boldsymbol{\omega}\right)
\right\vert^2
\nonumber\\
&\ge
\eta^2\left(\alpha\right)
\int_{\left\Vert\boldsymbol{\omega}\right\Vert_{\infty}\leq\alpha}
\left\vert
\mathcal{F}\left(\boldsymbol{\omega}\right)
-
\mathcal{G}\left(\boldsymbol{\omega}\right)
\right\vert^2
\nonumber\\
&\geq
\varepsilon^2
\eta^2\left(\alpha\right)
\left[
1-\zeta\left(\alpha^{-1}\right)
\right].
\nonumber
\end{align}
The above chain of inequalities hold for all $\alpha>C$, therefore we have:
\begin{align}
\left\Vert
\left(f-g\right)*Q
\right\Vert^2_2
&\ge
\frac{\varepsilon}{\left(2\pi\right)^K}
\sup_{\alpha>C}~
\eta\left(\alpha\right)
\sqrt{1-\zeta\left(\alpha^{-1}\right)},
\end{align}
which completes the proof.
\end{proof}


Theorem \ref{thm:generalResultTheorem-Appendix} presents a general approach to prove the recoverability of latent functions (or objects, which are the main focus in this work) from a certain class of independent additive noise. This approach works as long as the function class as well as the noise distribution are mostly comprised of low-frequency components in the Forier domain. For example, the Gaussian noise hurts low-frequency parts of a geometric object far less than its high-frequency details. More specifically, we prove the following corollary for Theorem \ref{thm:generalResultTheorem-Appendix}:
\begin{corollary}[Recoverability from Additive Gaussian Noise $\mathcal{N}\left(\boldsymbol{0},\sigma^2\boldsymbol{I}\right)$]
\label{corl:GaussinNoiseMain-Appendix}
Consider the setting in Theorem \ref{thm:generalResultTheorem-Appendix}, and assume the noise distribution follows $Q\triangleq\mathcal{N}\left(\boldsymbol{0},\sigma^2\boldsymbol{I}\right)$ for $\sigma>0$. Then, as long as for $f,g\in\mathscr{F}$ we have $\left\Vert f-g\right\Vert_2\ge\varepsilon$ for some $\varepsilon\ge0$, we also have
$$
\left\Vert
\left(f-g\right)*Q
\right\Vert_2
\ge
\frac{\varepsilon}{\left(2\pi\right)^K}
\left(
\sup_{\alpha>C}~
\sqrt{1-\zeta\left(\frac{1}{\alpha}\right)}
e^{-K\left(\sigma\alpha\right)^2/2}
\right)
$$
\end{corollary}
\begin{proof}
The Fourier transform of $Q=\mathcal{N}\left(\boldsymbol{0},\sigma^2\boldsymbol{I}\right)$ can be computed as follows:
\begin{equation}
\mathcal{F}\left\{Q\right\}\left(\boldsymbol{\omega}\right)
=
\prod_{i=1}^{K}\mathcal{F}\left\{\mathcal{N}\left(0,\sigma^2\right)\right\}\left(\omega_i\right)
=
e^{-\sigma^2\left\Vert\boldsymbol{\omega}\right\Vert_2^2/2}.
\end{equation}
Also, it can be easily checked that
$$
\inf_{\left\Vert\boldsymbol{\omega}\right\Vert_{\infty}\leq\alpha}
e^{-\sigma^2\left\Vert\boldsymbol{\omega}\right\Vert_2^2/2}=
e^{-\sigma^2/2\left(\alpha^2+\ldots+\alpha^2\right)}
=
e^{-K\left(\alpha\sigma\right)^2/2}.
$$
By subsititution into the end result of Theorem \ref{thm:generalResultTheorem-Appendix}, the claimed bounds can be achieved and the proof is complete.
\end{proof}


In this regard, our main explicit theoretical contribution in this section with respect to simplices has been stated in the following theorem:

\begin{theorem}[Recoverability of Simplices from Additive Noise]
\label{thm:mainNoisySimplexBound-Appendix}
For any two $\left(\bar{\lambda},\underline{\lambda}\right)$-regular simplices $\mathcal{S}_1,\mathcal{S}_2\in\mathbb{S}_K$ with $\bar{\lambda},\underline{\lambda}>0$, given that
$$
\mathcal{D}_{\mathrm{TV}}\left(\mathbb{P}_{\mathcal{S}_1*G_{\sigma}},\mathbb{P}_{\mathcal{S}_2*G_{\sigma}}\right)
=
\frac{1}{2}\int\left\vert
\left(f_{\mathcal{S}_1}-f_{\mathcal{S}_2}\right)*G_{\sigma}
\right\vert
\leq\varepsilon
$$
for some $\varepsilon\ge0$, where $G_{\sigma}$ (for $\sigma\ge0$) represents the density function associated to a Gaussian measure with zero mean and covariance matrix of $\sigma^2\boldsymbol{I}_{K\times K}$ in $\mathbb{R}^K$, i.e., $\mathcal{N}\left(\boldsymbol{0},\sigma^2\boldsymbol{I}\right)$. Then, we have
$$
\mathcal{D}_{\mathrm{TV}}\left(\mathbb{P}_{\mathcal{S}_1},\mathbb{P}_{\mathcal{S}_2}\right)
\leq \varepsilon
e^{\Omega\left(\frac{K}{\mathrm{SNR}^2}\right)}
,
$$
where $\mathrm{SNR}\triangleq \frac{\bar{\lambda}}{K\sigma}$ denotes the effective signal-to-noise ratio, which is the ratio of the standard deviation of a scaled uniform Dirichlet distribution to that of the noise, per dimension.
\end{theorem}
\begin{proof}
Proof is based on properties of the Fourier transforms of simplices $\mathcal{S}_1$ and $\mathcal{S}_2$.
For $K\in\mathbb{N}$, and any integrable function $f:\mathbb{R}^K\rightarrow\mathbb{R}$, the Fourier transform of $f$, denoted by $\mathcal{F}\left\{f\right\}\left(\boldsymbol{\omega}\right):\mathbb{R}^K\rightarrow\mathbb{C}$ is defined as follows:
\begin{equation}
\label{eq:FourierDef}
\mathcal{F}\left\{f\right\}\left(\boldsymbol{\omega}\right)
\triangleq
\int_{\boldsymbol{x}\in\mathbb{R}^K}
f\left(\boldsymbol{x}\right)
e^{-i\boldsymbol{\omega}^T\boldsymbol{x}}
\mathrm{d}\boldsymbol{x}.
\end{equation}
Throughout this proof, for $\mathcal{S}\in\mathbb{S}_K$, let us denote by $\mathcal{F}_{\mathcal{S}}$ the Fourier transform of $f_{\mathcal{S}}$, i.e., the uniform proability density function over $\mathcal{S}$. Also, the inverse Fourier transform which recovers $f_{\mathcal{S}}$ from $\mathcal{F}_{\mathcal{S}}$ can be written as
$$
f_{\mathcal{S}}\left(\boldsymbol{x}\right)
=
\frac{1}{2\pi}\int_{\boldsymbol{\omega}\in\mathbb{R}^K}\mathcal{F}_{\mathcal{S}}\left(\boldsymbol{\omega}\right)
e^{i\boldsymbol{\omega}^T\boldsymbol{x}}
\mathrm{d}\boldsymbol{\omega}.
$$

In this regard, let $\Delta_K$ represent the standard simplex in $\mathbb{R}^K$ which means a $K$-simplex with $\boldsymbol{\theta}_0=\boldsymbol{0}$ and $\boldsymbol{\Theta}=\boldsymbol{I}$. We begin by deriving a number of useful properties for $\Delta_K$ through the following lemmas.


\begin{lemma}
\label{lemma:consistency:recursive}
Let $\mathscr{F}_{\Delta_k}\left(\omega_1,\ldots,\omega_k\right):\mathbb{R}^k\rightarrow\mathbb{C}$ for $k\in[K]$ represent the Fourier transform of $f_{\Delta_k}$ in $\mathbb{R}^k$. Also, we have $\boldsymbol{\omega}_{1:k}\triangleq\left(\omega_1,\ldots,\omega_k\right)$. Then, the following recursive relation holds for $k>1$:
$$
\mathscr{F}_{\Delta_k}\left(
\boldsymbol{\omega}_{1:k}
\right)=
\frac{k}{i\omega_k}\left[
\mathscr{F}_{\Delta_{k-1}}
\left(
\boldsymbol{\omega}_{1:k-1}
\right)
-
e^{-i\omega_k}
\mathscr{F}_{\Delta_{k-1}}
\left(\omega_1-\omega_k,\ldots,\omega_{k-1}-\omega_k\right)
\right].
$$
\end{lemma}
\begin{proof}
For any $\left(x_1,\ldots,x_k\right)\in\Delta_k$,
Let $Z_i\triangleq x_1+\ldots+x_i$ for $i\in[k]$. Then
\begin{align}
\mathscr{F}_{\Delta_k}\left(\boldsymbol{\omega}_{1:k}\right)
&=
\int_{0}^{1}\int_{0}^{1-Z_1}\cdots
\int_{0}^{1-Z_{k-1}}
k!
e^{-i\left(\omega_1 x_1+\ldots+\omega_k x_k\right)}
\mathrm{d}x_1\ldots\mathrm{d}x_k
\\
&=\int_{0}^{1}\cdots\int_{0}^{1-Z_{k-2}}
k!
e^{-i\left(\omega_1 x_1+\ldots+\omega_{k-1} x_{k-1}\right)}
\left(
\int_{0}^{1-Z_{k-1}}
e^{-i\omega_k x_k}
\mathrm{d}x_k
\right)
\mathrm{d}x_1\ldots\mathrm{d}x_{k-1},
\nonumber
\end{align}
where a uniform probability density function over $\Delta_k$ has been assumed to be $f_{\Delta_k}\left(\boldsymbol{x}\right)=(k!)\boldsymbol{1}\left(\boldsymbol{x}\in\Delta_k\right)$ for all $\boldsymbol{x}\in\mathbb{R}^k$, due to the fact the Lebesgue measure (or volume) of the standard simplex is $1/k!$.
Since we have
$$
\int_{0}^{1-Z_{k-1}}
e^{-i\omega_k x_k}
\mathrm{d}x_k
=\frac{1}{i\omega_k}\left(
1-e^{-i\omega_{k}\left(1-Z_{k-1}\right)}
\right),
$$
the Fourier transform can be rewritten as follows:
\begin{align}
\mathscr{F}_{\Delta_k}\left(\boldsymbol{\omega}_{1:k}\right)
=&
\frac{k}{i\omega_k}
\int_{0}^{1}\cdots\int_{0}^{1-Z_{k-2}}
\left(k-1\right)!
e^{-i\left(\omega_1 x_1+\ldots+\omega_{k-1} x_{k-1}\right)}
\mathrm{d}x_1\ldots\mathrm{d}x_{k-1}
\nonumber\\
&-\frac{ke^{-i\omega_k}}{i\omega_k}
\int_{0}^{1}\cdots\int_{0}^{1-Z_{k-2}}
\left(k-1\right)!
\left(\prod_{i=1}^{k-1}e^{-i\left(\omega_i-\omega_k\right)x_i}
\right)
\mathrm{d}x_1\ldots\mathrm{d}x_{k-1}
\nonumber\\
=&
\frac{k}{i\omega_k}\left[
\mathscr{F}_{\Delta_{k-1}}
\left(
\boldsymbol{\omega}_{1:k-1}
\right)
-
e^{-i\omega_k}
\mathscr{F}_{\Delta_{k-1}}
\left(\omega_1-\omega_k,\ldots,\omega_{k-1}-\omega_k\right)
\right],
\end{align}
which completes the proof.
\end{proof}
Lemma \ref{lemma:consistency:recursive} gives us a recursive procedure to produce the Fourier transform, or derive usefull properties for $K$-dimensional standard simplex through tools such as induction. The following lemma establishes a relation between the Fourier transform of the standard simplex and that of an arbitrary simplex in $\mathbb{S}_K$ with a non-zero Lebesgue measure.

\begin{lemma}
\label{lemma:consistency:standard}
For a simplex $S\in\mathbb{S}_K$ with vertices $\boldsymbol{\theta}_0,\ldots,\boldsymbol{\theta}_K$ and a reversible zero-translated vertex matrix $\boldsymbol{\Theta}$, we have
$$
\mathscr{F}_{\mathcal{S}}\left(
\boldsymbol{\omega}_{1:k}
\right)
=
e^{-i\boldsymbol{\omega}_{1:k}^T\boldsymbol{\theta}_0}
\mathscr{F}_{\Delta_K}\left(
\boldsymbol{\Theta}^T
\boldsymbol{\omega}_{1:k}
\right).
$$
\end{lemma}
\begin{proof}
Let $f_{\mathcal{S}}:\mathbb{R}^K\rightarrow\mathbb{R}_{\ge0}$ denote the probability density function associated to $\mathcal{S}$. Then, it can be seen that
$$
f_{\mathcal{S}}\left(\boldsymbol{x}\right)
=
\frac{1}{\mathrm{det}\left(\boldsymbol{\Theta}\right)}
f_{\Delta_K}\left(
\boldsymbol{\Theta}^{-1}
\left(\boldsymbol{x}-\boldsymbol{\theta}_0\right)
\right),
\quad
\forall \boldsymbol{x}\in\mathbb{R}^K.
$$
In this regard, one just needs to write down the definition of Fourier transform for simplex $\mathcal{S}$ and utilizes the change of variables technique as follows:
\begin{align}
\mathscr{F}_{\mathcal{S}}\left(
\boldsymbol{\omega}_{1:k}
\right)
&=
\int_{\mathbb{R}^K}
f_{\mathcal{S}}\left(\boldsymbol{x}_{1:k}\right)
e^{-i\boldsymbol{\omega}_{1:k}^T\boldsymbol{x}_{1:k}}
\mathrm{d}x_1\ldots\mathrm{d}x_k
\\
&=
\frac{1}{\mathrm{det}\left(\boldsymbol{\Theta}\right)}
\int_{\mathbb{R}^K}
f_{\Delta_K}\left(
\boldsymbol{\Theta}^{-1}
\left(\boldsymbol{x}-\boldsymbol{\theta}_0\right)
\right)
e^{-i\boldsymbol{\omega}_{1:k}^T\boldsymbol{x}_{1:k}}
\mathrm{d}x_1\ldots\mathrm{d}x_k
\nonumber\\
&=
e^{-i\boldsymbol{\omega}_{1:k}^T\boldsymbol{\theta}_0}
\int_{\mathbb{R}^K}
f_{\Delta_K}\left(
\boldsymbol{u}_{1:k}
\right)
e^{-i\boldsymbol{\omega}_{1:k}^T
\boldsymbol{\Theta}\boldsymbol{u}_{1:k}}
\mathrm{d}u_1\ldots\mathrm{d}u_k
\nonumber\\
&=
e^{-i\boldsymbol{\omega}_{1:k}^T\boldsymbol{\theta}_0}
\mathscr{F}_{\Delta_K}\left(
\boldsymbol{\Theta}^T
\boldsymbol{\omega}_{1:k}
\right).
\end{align}
Therefore, the proof is complete.
\end{proof}


In the following lemma, we show that the uniform measure over the standard simplex $\Delta_K$ corresponds to a low-frequency probability density function. This would be the first step toward using Corollary \ref{corl:GaussinNoiseMain-Appendix}, in order to prove Theorem \ref{thm:mainNoisySimplexBound-Appendix}.

\begin{lemma}[Low-Pass Property of $\Delta_K$]
There exists a universal constant $C>0$, such that
for $\alpha>C$ and $K\in\mathbb{N}$ the uniform probability density function over $\Delta_K$, i.e., $f_{{\Delta_K}}:\mathbb{R}^K\rightarrow\mathbb{R}_{\ge0}$, is a low-frequency function in the following sense:
\begin{equation}
\frac{\mathrm{Vol}\left(\Delta_K\right)}{\left(2\pi\right)^K}
\int_{\left\Vert\boldsymbol{\omega}\right\Vert_{\infty}\ge\alpha}
\left\vert
\mathscr{F}_{\Delta_K}
\left(\boldsymbol{\omega}\right)
\right\vert^2
\leq
\mathcal{O}\left(\frac{K}{\alpha}\right),
\label{eq:simlexLowPass}
\end{equation}
where by $\left\Vert\boldsymbol{\omega}\right\Vert_{\infty}$ we simply mean $\max_{i\in[K]}\left\vert\omega_i\right\vert$.
\label{lemma:simplexLowPass}
\end{lemma}
\begin{proof}
Proof is based on the direct analysis of the Fourier transform of $f_{\Delta_k}$, which we denoted as $\mathscr{F}_{\Delta_K}\left(\boldsymbol{\omega}\right)$. We take advantage of the fact that the $K$-dimensional unit hypercube $\left[0,1\right]^K$ can be thought as the union of $K!$ properly rotated and translated versions of $\Delta_K$.

Mathematically speaking, assume the ordered tuple of unit axis-aligned vectors $E=\left(\boldsymbol{1}_1,\ldots,\boldsymbol{1}_K\right)$, where $\boldsymbol{1}_i$ for $i\in[K]$ denotes the one-hot vector over the $i$th component. Then,
let $\boldsymbol{V}_1,\ldots,\boldsymbol{V}_{K!}\in\mathbb{R}^{K\times K}$ be the set of orthonormal matrices, where each matrix transforms $E$ into one of its $K!$ possible permutations, i.e.,
$$
\left(\boldsymbol{V}_i\boldsymbol{1}_1,\ldots,\boldsymbol{V}_i\boldsymbol{1}_K\right)
=
\left(\boldsymbol{1}_{p_{i,1}},\ldots,\boldsymbol{1}_{p_{i,K}}\right),\quad
\boldsymbol{p}_i\in\mathrm{Perm}\left([K]\right).
$$
In this regard, we already know that there exist $K!$ corresponding vectors $\boldsymbol{b}_1,\ldots,\boldsymbol{b}_{K!}\in\mathbb{R}^K$, such that the following combined probability density function
\begin{equation}
\label{eq:lemma3:simp2cube}
\frac{1}{K!}\sum_{i=1}^{K!}
f_{\left(\boldsymbol{V}_i\left[\Delta_K+\boldsymbol{b}_i\right]\right)}
=f_{\square_K},
\end{equation}
where $f_{\square_K}$ denotes the pdf over the $K$-dimensional unit hypercube. Also, it should be noted that any two distinct summands in \eqref{eq:lemma3:simp2cube} have an empty overlap. In fact, each $\boldsymbol{V}_i\left[\Delta_K+\boldsymbol{b}_i\right]$ represents a translated and rotatated version of the standard simplex $\Delta_K$.

Next, one can see that
\begin{align}
\mathcal{F}\left\{
\frac{1}{K!}\sum_{i=1}^{K!}
f_{\left(\boldsymbol{V}_i\left[\Delta_K+\boldsymbol{b}_i\right]\right)}
\right\}
&=
\frac{1}{K!}\sum_{i=1}^{K!}
\mathcal{F}\left\{
f_{\left(\boldsymbol{V}_i\left[\Delta_K+\boldsymbol{b}_i\right]\right)}
\right\}
\nonumber\\
&=
\frac{1}{K!}\sum_{i=1}^{K!}
e^{-i\boldsymbol{\omega}^T\boldsymbol{b}_i}
\mathscr{F}_{\Delta_K}\left(\boldsymbol{V}^{-1}_i\boldsymbol{\omega}\right),
\end{align}
where we have used the result of Lemma \ref{lemma:consistency:standard}. Also, note that $\boldsymbol{V}^{-1}_i=\boldsymbol{V}_i$ for all $i$, and $\boldsymbol{V_i}\boldsymbol{\omega}$ represents a permutation of the components of $\boldsymbol{\omega}$. As a result and due to the symmtery of the standard simplex $\Delta_K$ and also the $\ell_{\infty}$-norm w.r.t. the ordering of the edges, we have
\begin{equation}
\int_{\left\Vert
\boldsymbol{\omega}
\right\Vert\leq\alpha
}
\left\vert
\mathcal{F}\left\{
f_{\left(\boldsymbol{V}_i\left[\Delta_K+\boldsymbol{b}_i\right]\right)}
\right\}
\left(\boldsymbol{\omega}\right)
\right\vert^2
=
\int_{\left\Vert
\boldsymbol{\omega}
\right\Vert\leq\alpha
}
\left\vert
\mathscr{F}_{\Delta_K}
\left(\boldsymbol{V}_i\boldsymbol{\omega}\right)
\right\vert^2
=
\int_{\left\Vert
\boldsymbol{\omega}
\right\Vert\leq\alpha
}
\left\vert
\mathscr{F}_{\Delta_K}
\left(\boldsymbol{\omega}\right)
\right\vert^2,\quad\forall i\in[K].
\end{equation}
Now, using \eqref{eq:lemma3:simp2cube} we have:
\begin{align}
\int_{\left\Vert
\boldsymbol{\omega}
\right\Vert\leq\alpha
}
\left\vert
\frac{1}{K!}\sum_{i=1}^{K!}
\mathcal{F}\left\{
f_{\left(\boldsymbol{V}_i\left[\Delta_K+\boldsymbol{b}_i\right]\right)}
\right\}
\right\vert^2
&=
\frac{1}{\left(K!\right)^2}
\sum_{i,j}^{K!}
\int_{\left\Vert
\boldsymbol{\omega}
\right\Vert\leq\alpha
}
\left\vert
\mathscr{F}_{\boldsymbol{V}^{-1}_i\Delta_K}
\bar{\mathscr{F}}_{\boldsymbol{V}^{-1}_j\Delta_K}
\right\vert
\left\vert
e^{-i\boldsymbol{\omega}^T\left(\boldsymbol{b}_i-\boldsymbol{b}_j\right)}
\right\vert
\nonumber\\
&=
\frac{1}{K!}
\int_{\left\Vert
\boldsymbol{\omega}
\right\Vert\leq\alpha
}
\left\vert
\mathscr{F}_{\Delta_K}\left(\boldsymbol{\omega}\right)
\right\vert^2
+\left(1-\frac{1}{K!}\right)\mathcal{O}\left(\frac{1}{\alpha}\right).
\label{eq:lemma3:firstO}
\end{align}
The latter term in the r.h.s. of \eqref{eq:lemma3:firstO} corresponds to the $\left(K!\right)^2-K!$ summands for which we have $i\neq j$. In fact, if $i\neq j$ and based on the fact that Fourier transform preserves inner product\footnote{This is a direct result of the fact that Fourier transform is an ``orthonormal" transformation.}, we have
$$
\int_{\mathbb{R}^K}\left\vert
\mathscr{F}_{\boldsymbol{V}_i\Delta_K}
\bar{\mathscr{F}}_{\boldsymbol{V}_j\Delta_K}
e^{-i\boldsymbol{\omega}^T\left(\boldsymbol{b}_i-\boldsymbol{b}_j\right)}
\right\vert
=
\int_{\mathbb{R}^K}
f_{\boldsymbol{V}_i\left[\Delta_K+\boldsymbol{b}_i\right]}
f_{\boldsymbol{V}_j\left[\Delta_K+\boldsymbol{b}_j\right]}
=0,
$$
which holds since $\Delta_i\triangleq\boldsymbol{V}_i\left[\Delta_K+\boldsymbol{b}_i\right]$ and $\Delta_j\triangleq\boldsymbol{V}_j\left[\Delta_K+\boldsymbol{b}_j\right]$ do not overlap with each other. However, when we add the constraint 
$\left\Vert\boldsymbol{\omega}\right\Vert_{\infty}\leq\alpha$, we effectively take the integral over the hypercube $\left[-\alpha,\alpha\right]^K$ instead of the whole $\mathbb{R}^K$. This procedure is equivalent to the innner product of the two ``smoothed" versions of $f_{\Delta_i}$ and $f_{\Delta_j}$ in the spatial domain. Here, by ``smoothed" we simply mean being convolved with a $K$-dimensional sinc function with parameter $\alpha$, i.e.,
$$
\mathrm{sinc}_{\alpha,K}\left(\boldsymbol{x}\right)
\triangleq
\prod_{i=1}^{K}\frac{\sin\left(\alpha x_i\right)}{\alpha x_i}.
$$
Hence, we have
\begin{align}
\int_{
\left\Vert\boldsymbol{\omega}\right\Vert_{\infty}\leq\alpha
}
\left\vert
\mathscr{F}_{\Delta_i}
\bar{\mathscr{F}}_{\Delta_j}
\right\vert
=
\int_{\mathbb{R}^K}
\left[
f_{\Delta_i}*
\mathrm{sinc}_{\alpha,K}
\right]\left[
f_{\Delta_j}*
\mathrm{sinc}_{\alpha,K}
\right],
\end{align}
which is at most $\mathcal{O}\left(1/\alpha\right)$, since only the leakages that are due to $\mathrm{sinc}_{\alpha,K}$ overlap with each other.

On the other hand, the l.h.s. of \eqref{eq:lemma3:firstO} represents the integration of the Fourier transform of $\square_K$ within $\left[-\alpha,\alpha\right]^K$. Therefore, we have
\begin{align}
\frac{1}{\left(2\pi\right)^KK!}
\int_{\left\Vert
\boldsymbol{\omega}
\right\Vert\leq\alpha
}
\left\vert
\mathscr{F}_{\Delta_K}\left(\boldsymbol{\omega}\right)
\right\vert^2
&\ge
\frac{1}{\left(2\pi\right)^K}
\int_{\left\Vert
\boldsymbol{\omega}
\right\Vert\leq\alpha
}
\left\vert
\mathscr{F}_{\square_K}\left(\boldsymbol{\omega}\right)
\right\vert^2
-
\mathcal{O}\left(\frac{1}{\alpha}\right)
\nonumber\\
&=
\prod_{i=1}^{K}
\frac{1}{2\pi}
\left(\int_{-\alpha}^{\alpha}
\left\vert
\int_{0}^{1}
e^{-i\omega_i x_i}
\right\vert^2
\right)
-
\mathcal{O}\left(\frac{1}{\alpha}\right)
\nonumber\\
&=
\left(1-\frac{2}{\pi\alpha}+o\left(\alpha^{-1}\right)\right)^K
-\mathcal{O}\left(\frac{1}{\alpha}\right)
\nonumber\\
&=1-\mathcal{O}\left(\frac{K}{\alpha}\right),
\end{align}
where we have used Laurent series expansion for the integral of sinc function to derive the bound. Finally, noting the fact that we have $\mathrm{Vol}\left(\Delta_K\right)=1/K!$, and also the Parseval's theorem:
$$
\frac{1}{\left(2\pi\right)^K}
\int_{\mathbb{R}^K}
\left\vert
\mathscr{F}_{\Delta_K}\left(\boldsymbol{\omega}\right)
\right\vert^2
=
\int_{\mathbb{R}^K}
f^2_{\Delta_K}\left(\boldsymbol{x}\right)
=\frac{\mathrm{Vol}\left(\Delta_K\right)}{\mathrm{Vol}^2\left(\Delta_K\right)}=
\mathrm{Vol}^{-1}\left(\Delta_K\right)
$$
completes the proof.
\end{proof}


So far, we have managed to show that the standard simplex $\Delta_K$ is associated to a low-frequency PDF and thus would preserve a minimum level of information even after getting corrupted by additive Gaussian noise. Using Lemma \ref{lemma:consistency:standard}, one can simply extend this notion to any $\left(\bar{\lambda},\underline{\lambda}\right)$-regular simplex. Before that, let us also extend this notion of ``low-frequency" property to the difference function
\begin{equation}
f_{\Delta_K}-f_{\boldsymbol{A}_{\varepsilon}\left(\Delta_K+\boldsymbol{b}_{\varepsilon}\right)},
\label{eq:lemma3:diffFunctionDef}
\end{equation}
where the linear transformation matrix $\boldsymbol{A}_{\varepsilon}$ and translation vector $\boldsymbol{b}_{\varepsilon}$ (for $\varepsilon>0$) are $\varepsilon$-controlled perturbations that alter the standard simplex with the following magnitude
$$
\mathcal{D}_{\mathrm{TV}}\left(
\mathbb{P}_{\Delta_K}
,
\mathbb{P}_{\boldsymbol{A}_{\varepsilon}\left(\Delta_K+\boldsymbol{b}_{\varepsilon}\right)}
\right)
=\varepsilon.
$$
This task is more challenging since the difference PDF function in \eqref{eq:lemma3:diffFunctionDef} can be both positive and negative, and in fact, integrates to zero over $\mathbb{R}^K$. However, the following lemma states that such functions for any $\boldsymbol{A}_{\varepsilon}$ and $\boldsymbol{b}_{\varepsilon}$ are ``band-pass" in nature: even though in the Fourier domain they become exactly zero at the origin, they also become infinitesimally small in terms of magnitude as one asymotitically increases $\left\Vert\boldsymbol{\omega}\right\Vert_{\infty}$.


\begin{lemma}[Low-Frequency Property for Difference of Simplices]
For $\varepsilon\ge0$, Assume the linear transformation matrix $\boldsymbol{A}_{\varepsilon}\in\mathbb{R}^{K\times K}$ where $A_{i,j}=\delta_{ij}+\mathcal{O}\left(\varepsilon\right)$, with $\delta_{ij}$ denoting the kroecker's delta function. Also, assume the translation vector $\boldsymbol{b}_{\varepsilon}\in\mathbb{R}^K$ where $b_i=\mathcal{O}\left(\varepsilon\right)$. Let $\hat{\Delta}_{K,\varepsilon}$ denote the simplex $\boldsymbol{A}_{\varepsilon}\left(\Delta_K+\boldsymbol{b}_K\right)$, and assume the total variation distance between $\mathbb{P}_{\Delta_K}$ and $\mathbb{P}_{\hat{\Delta}_{K,\varepsilon}}$ is $\varepsilon$. Then
$$
\frac{1}{\left(2\pi\right)^K}
\int_{\left\Vert
\boldsymbol{\omega}
\right\Vert_{\infty}\geq\alpha
}
\left\Vert
\mathscr{F}_{\Delta_K} - \mathscr{F}_{\hat{\Delta}_{K,\varepsilon}}
\right\Vert^2
\leq
\mathcal{O}\left(\frac{K\varepsilon}{\alpha}\right),
$$
for sufficiently large $\alpha>0$.
\label{lemma:simpDiffLowPass}
\end{lemma}
\begin{proof}
Based on Lemmas \ref{lemma:simplexLowPass} and \ref{lemma:consistency:standard} and for sufficiently small $\varepsilon$, we already know that both $\mathscr{F}_{\Delta_K}$ and $\mathscr{F}_{\hat{\Delta}_{K,\varepsilon}}$ are low-frequency functions in the following sense: their normalized $\ell_2$ energy outside of the hypercube $\left[-\alpha,\alpha\right]^K$ is at most $\mathcal{O}\left(K/\alpha\right)$. We also know that
$$
\frac{1}{\left(2\pi\right)^K}
\int_{\left\Vert
\boldsymbol{\omega}
\right\Vert_{\infty}\geq\alpha
}
\left\Vert
\mathscr{F}_{\Delta_K} - \mathscr{F}_{\hat{\Delta}_{K,\varepsilon}}
\right\Vert^2
=
\mathcal{O}\left(\varepsilon\right),
$$
due to the assumption of the lemma. However, in order to prove the $\leq\mathcal{O}\left(K\varepsilon/\alpha\right)$ bound, we need to prove another important property: that the energy distribution of the difference function is not significantly controlled by $\varepsilon$, e.g., it does not concentrate on high-frequency regions with $\left\Vert\boldsymbol{\omega}\right\Vert_{\infty}\ge\mathcal{O}\left(1/\varepsilon\right)$.

Before going further into the mathematical details of the above statement, let us present a better view on the Fourier transform of a simplex. We start this procedure with computing $\mathscr{F}_{\Delta_K}$. In this regard, we show the following relation holds for $k\in\mathbb{N}$:
\begin{equation}
\label{eq:pichideh}
\mathcal{F}_{\Delta_k}
=
\frac{k!}{i^k}
\sum_{\ell=1}^{k}\left(1-e^{-i\omega_\ell}\right)
\left[
\omega_\ell
\prod_{j\neq\ell}^{k}\left(\omega_j-\omega_\ell\right)
\right]^{-1}.
\end{equation}
Proof of \eqref{eq:pichideh} is by induction. {\bf {Base}}: First, we show the equality holds for the simple case of $k=1$. {\bf {Step}}: Then, we prove that for $k\ge 2$, if the relation holds for $k-1$, then it also holds for $k$.

\paragraph{Base}
For $k=1$, it can be readily seen that a uniform measure over $\Delta_1$ is a {\it {unit~pulse~function}} over the interval $[0,1]$. Therefore, the Fourier transform of $f_{\Delta_1}$ is already known to be a sinc function, i.e.,
\begin{align}
\mathcal{F}_{\Delta_1}\left(\omega\right)=
\int_{0}^{1}e^{-i\omega x}
\mathrm{d}x
=
\frac{1-e^{-i\omega}}{i\omega},
\end{align}
which matches the formulation of \eqref{eq:pichideh} if one sets $k=1$.

\paragraph{Induction~step}

Assume \eqref{eq:pichideh} holds for $k-1$. By taking advantage of Lemma \ref{lemma:consistency:recursive}, we have:
\begin{align}
\mathcal{F}_{\Delta_k}
=&
\frac{k}{i\omega_k}
\left(
\mathcal{F}_{\Delta_{k-1}}\left(\omega_1,\ldots,\omega_{k-1}\right) - e^{-i\omega_k}
\mathcal{F}_{\Delta_{k-1}}\left(\omega_1-\omega_k,\ldots,\omega_{k-1}-\omega_k\right)
\right)
\\
=&
\frac{\left(k-1\right)!}{i^{k-1}}
\frac{k}{i\omega_k}
\left(
\sum_{\ell=1}^{k-1}\left(1-e^{-i\omega_\ell}\right)
\left[
\omega_\ell
\prod_{j\neq\ell}^{k-1}\left(\omega_j-\omega_\ell\right)
\right]^{-1}
\right.
\nonumber
\\
&
\hspace*{22mm}
-\left.
e^{-i\omega_k}
\sum_{\ell=1}^{k-1}
\left(1-e^{-i\left(\omega_\ell-\omega_k\right)}\right)
\left[
\left(\omega_\ell-\omega_k\right)
\prod_{j\neq\ell}^{k-1}\left(\omega_j-\omega_\ell\right)
\right]^{-1}
\right)
\nonumber
\\
=&
\frac{k!}{i^k}\sum_{\ell=1}^{k-1}
\left(
\left(\omega_k - \omega_\ell\right)
\left(
1-e^{-i\omega_\ell}
\right)
-
\omega_\ell
\left(
e^{-i\omega_\ell}-
e^{-i\omega_k}\right)
\right)
\left[
\omega_\ell
\omega_k
\prod_{j\neq\ell}^{k-1}\left(\omega_j-\omega_\ell\right)
\left(\omega_k-\omega_\ell\right)
\right]^{-1}
\nonumber
\\
=&
\frac{k!}{i^k}
\sum_{\ell=1}^{k-1}
\left(
\omega_k
\left(
1-
e^{-i\omega_\ell}\right)
-
\omega_\ell
\left(
1-e^{-i\omega_k}
\right)
\right)
\left[
\omega_\ell
\omega_k
\prod_{j\neq\ell}^{k-1}\left(\omega_j-\omega_\ell\right)
\left(\omega_k-\omega_\ell\right)
\right]^{-1}
\nonumber\\
=&
\frac{k!}{i^k}
\sum_{\ell=1}^{k-1}
\left(
1-e^{-i\omega_\ell}
\right)
\left[
\omega_\ell
\prod_{j\neq\ell}^{k}\left(\omega_j-\omega_\ell\right)
\right]^{-1}
-
\frac{k!}{i^k}
\left(
\frac{
1-e^{-i\omega_k}
}{\omega_k}
\right)
\sum_{\ell=1}^{k-1}
\prod_{j\neq\ell}^{k}\frac{1}{\omega_j-\omega_\ell}
\nonumber\\
=&
\frac{k!}{i^k}
\sum_{\ell=1}^{k}
\left(
\frac{1-e^{-i\omega_\ell}}{\omega_\ell}
\right)
\prod_{j\neq\ell}^{k}\frac{1}{\omega_j-\omega_\ell},
\nonumber
\end{align}
which again matches with the formulation of \eqref{eq:pichideh} and therefore completes the induction step. Here, for the last equality we have used the following mathematical identity:
$$
\sum_{\ell=1}^{k}\prod_{j\neq\ell}^{k}\frac{1}{\omega_j-\omega_\ell}=0.
$$
\hfill{\bf {End of induction}}
\\[2mm]
‌Based on this explicit formula, the difference function $\mathscr{F}_{\Delta_K} - \mathscr{F}_{\hat{\Delta}_{K,\varepsilon}}$ can be written as follows:
\begin{align}
\mathscr{F}_{\Delta_K} - \mathscr{F}_{\hat{\Delta}_{K,\varepsilon}}
&=
\frac{K!}{i^K}
\sum_{\ell=1}^{K}\left(1-e^{-i\omega_\ell}\right)
\left[
\omega_\ell
\prod_{j\neq\ell}^{K}\left(\omega_j-\omega_\ell\right)
\right]^{-1}
\left(1-r_{\ell}\left(\boldsymbol{\omega}\right)\right),
\end{align}
where
\begin{align}
r_{\ell}\left(\boldsymbol{\omega}\right)
&\triangleq
e^{-i\boldsymbol{\omega}^T\boldsymbol{b}_{\varepsilon}}
\left(
\frac{1-e^{-i\boldsymbol{\omega}^T\boldsymbol{a}_{\ell}}}{1-e^{-i\omega_{\ell}}}
\right)
\frac{\omega_{\ell}}{\boldsymbol{\omega}^T\boldsymbol{a}_{\ell}}
\prod_{j\neq \ell}
\frac{\omega_j-\omega_{\ell}}{\boldsymbol{\omega}^T\left(\boldsymbol{a}_j-\boldsymbol{a}_{\ell}\right)}
\nonumber\\
&=e^{-i\boldsymbol{\omega}^T
\left(\boldsymbol{b}_{\varepsilon}-\boldsymbol{\delta}_{\ell}\right)}
\frac{\sin\left(\omega_{\ell}/2 + \boldsymbol{\delta}^T_{\ell}\boldsymbol{\omega}/2\right)}
{\sin\left(\omega_{\ell}/2\right)}
\frac{1}{1+\left(\boldsymbol{\omega}/\omega_{\ell}\right)^T\boldsymbol{\delta}_{\ell}}
\prod_{j\neq \ell}
\frac{1}{1+
\frac{\left(\boldsymbol{\delta}_j-\boldsymbol{\delta}_{\ell}\right)^T\boldsymbol{\omega}
}{\left(\omega_j-\omega_\ell\right)}}
,
\end{align}
where $\boldsymbol{a}_i$ denotes the $i$th row of $\boldsymbol{A}_{\varepsilon}$. Also, we have assumed $\boldsymbol{a}_i=\boldsymbol{1}_i+\boldsymbol{\delta}_i$ with $\left\Vert\boldsymbol{\delta}_i\right\Vert_{\infty},~i\in[K]$ being upper bounded by $\mathcal{O}\left(\varepsilon\right)$ according to the definition of $\boldsymbol{A}_{\varepsilon}$. It can be readily checked that $r_{\ell}\left(\boldsymbol{\omega}\right)=1+\mathcal{O}\left(\varepsilon\right)$. Also, the multiplicative factors that form $r_{\ell}\left(\boldsymbol{\omega}\right)$ are either not dependent on $\left\Vert\boldsymbol{\omega}\right\Vert_2$ (dependence is only on the direction, and not the magnitude) or they oscillate as a function of magnitude. This implies that $r_{\ell}\left(\boldsymbol{\omega}\right)$ which appears solely due to the difference between the two simplices does not behave differently in regions which are far from the origin or the areas in its vicinity.

Due to the above derivations, the difference function has a similar behaviour to that of $\mathscr{F}_{\Delta_K}\left(\boldsymbol{\omega}\right)$ and thus we have
$$
\frac{1}{\left(2\pi\right)^K}
\int_{\left\Vert
\boldsymbol{\omega}
\right\Vert_{\infty}\geq\alpha
}
\left\Vert
\mathscr{F}_{\Delta_K} - \mathscr{F}_{\hat{\Delta}_{K,\varepsilon}}
\right\Vert^2
\leq
\mathcal{O}\left(\frac{K\varepsilon}{\alpha}\right),
$$
which completes the proof.
\end{proof}
With the help of Corollary \ref{corl:GaussinNoiseMain-Appendix} and Lemmas \ref{lemma:consistency:standard}, \ref{lemma:simplexLowPass} and \ref{lemma:simpDiffLowPass}, we can finally prove the claimed bound. Let simplex $\mathcal{S}_1$ and $\mathcal{S}_2$ to have the vertex matrices $\boldsymbol{\Theta}_1$ and $\boldsymbol{\Theta}_2$, respectively. Also, we have already assumed
$$
\underline{\lambda}\leq\lambda_{\min}\left(\boldsymbol{\Theta}_i\right)\quad,\quad
\lambda_{\max}\left(\boldsymbol{\Theta}_i\right)\leq\bar{\lambda}\quad
\mathrm{for}~i=1,2.
$$
Then, based on Lemma \ref{lemma:consistency:standard} and for a sufficiently large $\alpha>0$, we have
\begin{align}
\int_{\left\Vert\boldsymbol{\omega}\right\Vert_{\infty}\geq\alpha}
\left\vert
\mathscr{F}_{\mathcal{S}_i}\left(\boldsymbol{\omega}\right)
\right\vert^2
&=
\int_{\left\Vert\boldsymbol{\omega}\right\Vert_{\infty}\geq\alpha}
\left\vert
\mathscr{F}_{\Delta_K}\left(\boldsymbol{\Theta}_i^T\boldsymbol{\omega}\right)
\right\vert^2
\nonumber\\
&\leq
\frac{1}{\mathrm{det}\left(\boldsymbol{\Theta}\right)}
\int_{\left\Vert\boldsymbol{\omega}\right\Vert_{\infty}\geq\bar{\lambda}\alpha}
\left\vert
\mathscr{F}_{\Delta_K}\left(\boldsymbol{\omega}\right)
\right\vert^2
\nonumber\\
&\leq
\frac{\left(2\pi\right)^K}{\mathrm{Vol}\left(\mathcal{S}_i\right)}
\mathcal{O}\left(\frac{K}{\bar{\lambda}\alpha}\right),
\end{align}
for $i=1,2$. Therefore, Lemma \ref{lemma:simpDiffLowPass} would consequently imply the following relation for the difference function of the two simplices:
\begin{align}
\frac{1}{\left(2\pi\right)^K}
\int_{\left\Vert\boldsymbol{\omega}\right\Vert_{\infty}\geq\alpha}
\left\vert
\mathscr{F}_{\mathcal{S}_1}-\mathscr{F}_{\mathcal{S}_2}
\right\vert^2
\leq
\mathcal{O}\left(\frac{K}{\bar{\lambda}\alpha}\right)
\int_{\mathbb{R}^K}\left(
f_{\mathcal{S}_1}-f_{\mathcal{S}_2}
\right)^2,
\end{align}
which facilitates the usage of our main general result in Theorem \ref{thm:generalResultTheorem-Appendix}, and more specifically Corollary \ref{corl:GaussinNoiseMain-Appendix} which leads to the following relation:
\begin{align}
\left\Vert
\left(f_{\mathcal{S}_1}-f_{\mathcal{S}_2}\right)*G_{\sigma}
\right\Vert_2
&\ge 
\left\Vert f_{\mathcal{S}_1}-f_{\mathcal{S}_2} \right\Vert_2
\sup_{\alpha}
\left(
\sqrt{1-\mathcal{O}\left(\frac{K}{\bar{\lambda}\alpha}\right)}
e^{-K\left(\sigma\alpha\right)^2}
\right)
\nonumber\\
&\geq
\left\Vert f_{\mathcal{S}_1}-f_{\mathcal{S}_2} \right\Vert_2
e^{-\mathcal{O}\left(\frac{K}{\mathrm{SNR}^2}\right)},
\label{eq:theoremNoisy:mainfinal}
\end{align}
since $\alpha\geq\Omega\left(K/\Bar{\lambda}\right)$ guarantees that the l.h.s. of the bound remains positive, while gives the minimum possible exponent (at least order-wise) to the exponential term. Combining the fact $\ell_2$-norm $\leq$ $\ell_1$-norm together with theorem's assumptions, we have
$$
\left\Vert
\left(f_{\mathcal{S}_1}-f_{\mathcal{S}_2}\right)*G_{\sigma}
\right\Vert_2
\leq
\left\Vert
\left(f_{\mathcal{S}_1}-f_{\mathcal{S}_2}\right)*G_{\sigma}
\right\Vert_1
\leq 2\varepsilon.
$$
Consequently, \eqref{eq:theoremNoisy:mainfinal} implies the following upper-bound on the $\ell_2$-norm of the difference PDF function $f_{\mathcal{S}_1}-f_{\mathcal{S}_2}$ as follows:
$$
\left\Vert f_{\mathcal{S}_1}-f_{\mathcal{S}_2}\right\Vert_2\leq
\varepsilon e^{\Omega\left(\frac{K}{\mathrm{SNR}^2}\right)},
$$
and completes the proof.
\end{proof}

\end{document}